%% file: main.tex
\documentclass[journal]{IEEEtran}
\usepackage[ruled,vlined,linesnumbered,noresetcount]{algorithm2e}
\usepackage{booktabs} 
\usepackage{amsfonts}
\usepackage{amssymb}
\usepackage{mathrsfs}
\usepackage{comment}
\usepackage{bbm}
\usepackage{fouriernc} 
\usepackage[dvipsnames]{xcolor}
\usepackage{pgfplots}
\usepackage{subfig}
\usepackage{multirow}
\usepackage{dsfont}
\usepackage{upgreek}
\usepackage{siunitx}
\usepackage{mathcomp}
\usepackage{tikz-network}
\usepackage{romannum}
\usepackage{tikz-cd}
\usepackage{tikz}
\usepackage{lipsum}
\usepackage{multicol}
\usepackage{neuralnetwork}
\usepackage{ifpdf}
\usepackage{rotating}
\usepackage{booktabs} 
\usepackage{amsmath} 
\usepackage{amsfonts}
\usepackage{amssymb}
\usepackage{mathrsfs}
\usepackage{commath}
\usepackage[acronym]{glossaries}
\usepackage{ifpdf}
\usepackage{rotating}
\usepackage{cite}
\usepackage{tikzpagenodes}
\usepackage{pstricks}
\usepackage{pstricks-add}
\usepackage{calc}
\usepackage{alphalph}
\usepackage[hidelinks]{hyperref}
\usepackage{algorithm2e}
\usepackage{lscape}
\usepackage{mathtools}
\usepackage{enumerate}
\usepackage{scalerel}
\usetikzlibrary{positioning}
\definecolor{mygray}{gray}{0.6}
\definecolor{airforceblue}{rgb}{0.36, 0.54, 0.66}
\definecolor{applegreen}{rgb}{0.55, 0.71, 0.0}
\definecolor{orcidlogocol}{HTML}{A6CE39}
\newcommand{\tran}{^{\top\kern-\scriptspace}}
\usetikzlibrary{matrix,chains,positioning,decorations.pathreplacing,arrows}

\DeclareMathAlphabet{\pazocal}{OMS}{zplm}{m}{n}

\newcommand{\Lb}{\pazocal{L}}
\newcommand{\Db}{\pazocal{D}}

\usetikzlibrary{svg.path}

\definecolor{orcidlogocol}{HTML}{A6CE39}
\tikzset{
  orcidlogo/.pic={
    \fill[orcidlogocol] svg{M256,128c0,70.7-57.3,128-128,128C57.3,256,0,198.7,0,128C0,57.3,57.3,0,128,0C198.7,0,256,57.3,256,128z};
    \fill[white] svg{M86.3,186.2H70.9V79.1h15.4v48.4V186.2z}
                 svg{M108.9,79.1h41.6c39.6,0,57,28.3,57,53.6c0,27.5-21.5,53.6-56.8,53.6h-41.8V79.1z M124.3,172.4h24.5c34.9,0,42.9-26.5,42.9-39.7c0-21.5-13.7-39.7-43.7-39.7h-23.7V172.4z}
                 svg{M88.7,56.8c0,5.5-4.5,10.1-10.1,10.1c-5.6,0-10.1-4.6-10.1-10.1c0-5.6,4.5-10.1,10.1-10.1C84.2,46.7,88.7,51.3,88.7,56.8z};
  }
}

\newcommand\orcidicon[1]{\href{https://orcid.org/#1}{\mbox{\scalerel*{
\begin{tikzpicture}[yscale=-1,transform shape]
\pic{orcidlogo};
\end{tikzpicture}
}{|}}}}

\pgfplotsset{compat=1.17}
\begin{document}
\pagenumbering{arabic}
\title{Preference Neural Network}
\author{Ayman Elgharabawy\IEEEauthorrefmark{1}\orcidicon{0000-0003-3381-4708}\,, Mukesh Prasad \IEEEauthorrefmark{2}\orcidicon{0000-0002-7745-9667}\, Senior Member,
\textit{IEEE}, Chin-Teng Lin \IEEEauthorrefmark{3}\orcidicon{0000-0001-8371-8197}\,, Fellow, \textit{IEEE}%
\\


Email: \IEEEauthorrefmark{1}Ayman.Elgharabawy@anu.edu.au,
\IEEEauthorrefmark{2}Mukesh.Prasad@uts.edu.au,
\IEEEauthorrefmark{3}Chin-teng.lin@uts.edu.au\\
\IEEEauthorrefmark{1}Corresponding author
}
\markboth{IEEE TRANSACTIONS ON EMERGING TOPICS IN COMPUTATIONAL INTELLIGENCE}%
{Shell \MakeLowercase{\textit{et al.}}: Bare Demo of IEEEtran.cls for Journals}

\IEEEoverridecommandlockouts
\IEEEpubid{\makebox[\columnwidth]{978-1-5386-5541-2/18/\$31.00~\copyright2018 IEEE \hfill} \hspace{\columnsep}\makebox[\columnwidth]{ }}
\maketitle
\begin{abstract}
This paper proposes a novel label ranker network to learn the relationship between labels to solve ranking and classification problems. The Preference Neural Network (\textit{PNN}) uses \textit{spearman} correlation gradient ascent and two new activation functions, positive smooth staircase (\textit{PSS}), and smooth staircase (\textit{SS}) that accelerate the ranking by creating almost deterministic preference values. \textit{PNN} is proposed in two forms, fully connected simple Three layers and Preference Net (\textit{PN}), where the latter is the deep ranking form of \textit{PNN} to learning feature selection using ranking to solve images classification problem. \textit{PN} uses a new type of ranker kernel to generate a feature map. \textit{PNN} outperforms five previously proposed methods for label ranking, obtaining state-of-the-art results on label ranking, and \textit{PN} achieves promising results on \textit{CFAR-100} with high computational efficiency.
\end{abstract}

\section{Introduction}
\IEEEPARstart{P}REFERENCE learning (\textit{PL}) is an extended paradigm in machine learning that induces predictive preference models from experimental data~\cite{PLbook,prafman,adom}. \textit{PL} has applications in various research areas such as knowledge discovery and recommender systems~\cite{recommender}. Objects, instances, and label ranking are the three main categories of \textit{PL} domain. Of those, label ranking (\textit{LR}) is a challenging problem that has gained importance in information retrieval by search engines~\cite{aio,crammer}. Unlike the common problems of regression and classification~\cite{9714196,10.1093/comjnl/bxaa168,9750402,10004751,GUO202283,electronics11193022,9983500}, label ranking involves predicting the relationship between multiple label orders. For a given instance $\textit{x}$ from the instance space $\mathbbm{x}$, there is a label $\Lb$ associated with $\textit{x}$, $\Lb\in\pi$, where $\pi=\{\lambda_{1},..,\lambda_{n}$\}, and \textit{n} is the number of labels. \textit{LR} is an extension of multi-class and multi-label classification, where each instance $\textit{x}$ is assigned an ordering of all the class labels in the set $\Lb$. This ordering gives the ranking of the labels for the given $\textit{x}$ object. This ordering can be represented by a permutation set $\pi=\{1, 2, \cdots, n\}$. The label order has the following three features. irreflexive where $\lambda_{a}\nsucc\lambda_{a}$ ,transitive where ($\lambda_{a}\succ\lambda_{b})\land(\lambda_{b}\succ\lambda_{c}$) $\implies\lambda_{a}\succ\lambda_{c}$ and asymmetric $\lambda_{a}\succ\lambda_{b}\implies\lambda_{b}\nsucc\lambda_{a}$.
Label preference takes one of two forms, strict and non-strict order. The strict label order ($\lambda_{a}\succ\lambda_{b}\succ\lambda_{c}\succ\lambda_{d}$) can be represented as $\pi=(1, 2, 3, 4)$ and for non-restricted total order $\pi=(\lambda_{a}\succ\lambda_{b}\simeq\lambda_{c}\succ\lambda_{d})$ can be represented as $\pi=(1, 2, 2, 3)$, where $a, b, c, and, d$ are the label indexes and $\lambda_{a},\lambda_{b},\lambda_{c}$ and $\lambda_{d}$ are the ranking values of these labels.

For the non-continuous permutation space, The order is represented by the relations mentioned earlier and the $\bot$ incomparability binary relation. For example the partial order $\lambda_{a}\succ\lambda_{b}\succ\lambda_{d}$ can be represented as $\pi=(1,2,0,3)$ where 0 represents an incomparable relation since $\lambda_{c}$ is not comparable to ($\lambda_{a},\lambda_{b},\lambda_{d}$).

Various label ranking methods have been introduced in recent years~\cite{taxonamy}, such as decomposition-based methods, statistical methods, similarity, and ensemble-based methods. Decomposition methods include pairwise comparison~\cite{Furnkranz1,Furnkranz2}, log-linear models and constraint classification~\cite{constraint}.
The pairwise approach introduced by H{\"u}llermeier~\cite{Hullermeier} divides the label ranking problem into several binary classification problems to predict the pairs of labels $\lambda_{i}\succ\lambda_{j}$ or $\lambda_{j}\prec\lambda_{i}$ for an input \textit{x}.
 Statistical methods includes decision trees~\cite{dt1}, instance-based methods (Plackett-Luce)~\cite{plackett} and Gaussian mixture model based approaches. For example, Mihajlo uses Gaussian mixture models to learn soft pairwise label preferences~\cite{gaussian}.

The artificial neural network (\textit{ANN}) for ranking was first introduced as (RankNet) by Burge to solve the problem of object ranking for sorting web documents by a search engine~\cite{Burges}. Rank net uses gradient descent and probabilistic ranking cost function for each object pair. The multilayer perceptron for label ranking (\textit{MLP-LR})~\cite{Ribeiro} employs a network architecture using a \textit{sigmoid} activation function to calculate the error between the actual and expected values of the output labels. However, It uses a local approach to minimize the individual error per output neuron by subtracting the actual-predicted value and using Kendall error as a global approach. Neither direction uses a ranking objective function in backpropagation (\textit{BP}) or learning steps.

The deep neural network (\textit{DNN}) is introduced for object ranking to solve document retrieval problems. RankNet~\cite{Burges}, RankBoost~\cite{Freund}, and Lambda MART~\cite{svore}, and deep pairwise label ranking models ~\cite{deeppairwise}, are convolution neural Network (\textit{CNN}) approaches for the vector representation of the query and document-based.
\textit{CNN} is used for image retrieval~\cite{Deeplabelraning} and label classification for remote sensing and medical diagnosing~\cite{labelclass,Cherian2021ClassificationOR,singh,9137336,10.1145/3468506,9999261,https://doi.org/10.1029/2022WR033241,rs13224604}.
A multi-valued activation function has been proposed by Moraga and Heider~\cite{moraga} to propose a Generalized Multiple–valued Neuron with a differentiable soft staircase activation function, which is represented by a sum of a set of sigmoidal functions. In addition, Aizenberg proposed a generalized multiple-valued neuron using a convex shape to support complex numbers neural network and multi-values numbers~\cite{mvn1}.
Visual saliency detection using the Markov chain model is one approach that simulates the human visual system by highlighting the most important area in an image and calculating superpixels as absorbing nodes~\cite{8941002,10018569,https://doi.org/10.1049/cit2.12174}. However, this approach needs a saliency optimization on the results and has calculation cost~\cite{Jiang2020RobustVS},\cite{Gupta2020SalientOD}.

 Particle Swarm Optimization in movement detection is based on the concept of variation and inter-frame difference for feature selection. The swarm algorithms are mainly used in human motion detection in sports, and it is used based on probabilistic optimization algorithm~\cite{Lei2021SportsID,ZHANG2022109766,app13053082,9350239} and CNN~\cite{Zhang2022SportsAR}.

Some of the methods mentioned above and their variants have some issues that can be broadly categorized into three types: 
\begin{itemize}
\item[1)] The \textit{ANN} Predictive probability can be enhanced by limiting the output ranking values in the SS functions to a discrete value instead of a range of values of the rectified linear unit (\textit{Relu}), \textit{Sigmoid}, or \textit{Softmax} activation functions. The predictive is enhanced by using the \textit{SS} function slope as a step function to create discrete values, accelerating the learning by reducing the output values to accelerate the ranking convergence.

\item[2)] The drawback of ranking based on the classification technique ignores the relation between multiple labels: When the ranking model is constructed using binary classification models, these methods cannot consider the relationship between labels because the activation functions do not provide deterministic multiple values. Such ranking based on minimizing pairwise classification errors differs from maximizing the label ranking's performance considering all labels. This is because pairs have multiple models that may reduce ranking unification by increasing ranking pairs conflicts where there is no ground truth, which has no generalized model to rank all the labels simultaneously. For example, $\Db = (1,1,1)$ for $\pi=(\lambda_{a}\succ\lambda_{b}\succ\lambda_{c})$ and $\Db = (1,1,1)$ for $\pi=(\lambda_{a}\succ\lambda_{c}\succ\lambda_{b})$ the ranking is unique; however, pairwise classification creates no ground truth ranking for the pair $\lambda_{b}\succ\lambda_{c}$ and $\lambda_{c}\succ\lambda_{b}$ which adds more complexity to the learning process.

\item[3)] Ignoring the relation between features. The convolution kernel has a fixed size that detects one feature per kernel. Thus, it ignores the relationship between different parts of the image. For example, \textit{CNN} detects the face by combining features (the mouth, two eyes, the face oval, and a nose) with a high probability of classifying the subject without learning the relationship between these features. For example, the proposed \textit{PN} kernel start attention to the important features that have a high number of pixel ranking variation. 
\end{itemize}
The main contribution of the proposed neural network is 
\begin{itemize}
\item Solving the label ranking as a machine learning problem.
\item Solving the deep learning classification problem by employing computational ranking in feature selection and learning.
\end{itemize}

 Where \textit{PNN} has several advantages over existing label ranking methods and \textit{CNN} classification approaches.
\begin{itemize}
\item[1)]\textit{PNN} uses the smooth staircase \textit{SS} as an activation function that enhances the predictive probability over the \textit{sigmoid} and \textit{Softmax} due to the step shape that enhances the predictive probability from a range from -1 to 1 in the sigmoid to almost discrete multi-values.
\item[2)] \textit{PNN} uses gradient ascent to maximize the \textit{spearman} ranking correlation coefficient. In contrast, other classification-based methods such as \textit{MLP-LR} use the absolute difference of root mean square error (\textit{RMS}) by calculating the differences between actual and predicted ranking and other \textit{RMS} optimization, which may not give the best ranking results.
\item[3)]\textit{PNN} is implemented directly as a label ranker. It uses staircase activation functions to rank all the labels together in one model. The \textit{SS} or \textit{PSS} functions provide multiple output values during the conversions; however, \textit{MLP-LR} and \textit{RankNet} use \textit{sigmoid} and \textit{Relu} activation functions. These activation functions have a binary output. Thus, it ranks all the labels together in one model instead of pairwise ranking by classification.
\item[4)]\textit{PN} uses a novel approach for learning the feature selection by ranking the pixels and using different sizes of weighted kernels to scan the image and generate the features map. 

\end{itemize}
The next section explains the Ranker network experiment, problem formulation, and the \textit{PNN} components (Activation functions, Objective function, and network structure) that solve the Ranker problems and comparison between Ranker network and \textit{PNN}.

\section{\textit{PNN} Components}
\subsection{Initial Ranker}
The proposed \textit{PNN} is based on an initial experiment to implement a computationally efficient label ranker network based on the Kendall $\tau$ error function and \textit{sigmoid} activation function using simple structure as illustrated in section \Romannum{4} Fig.~\ref{fig:typea-1}.

 The ranker network is a fully connected, three-layer net. The input represents one instance of data with three inputs, and there are six neurons in the hidden layer and three output neurons representing the labels' index. Each neuron represents the ranking value. A small toy data set is used in this experiment. The ranker uses \textit{RMS} gradient descent as an error function to measure the difference between the predicted and actual ranking values. The ranker has Kendall $\tau$ as a stopping criterion.
 The same \textit{ANN} structure, number of neurons and learning rate using \textit{SS} activation function, and \textit{spearman} error function and gradient ascent of $\rho$ will be discussed in section \Romannum{4}.
 The ranking convergence reaches $\tau\simeq1$ after 160 epochs using the \textit{Sigmoid} function~\cite{videofile}.
 The \textit{sigmoid} and \textit{ReLU} shapes have a slightly high rate of change of $y$, and it produces a larger output range of data. Therefore, we consider ranking performance as one of the disadvantages of \textit{sigmoid} function in the ranker network. 

 The ranker network has two main problems. 
 \begin{itemize}
\item[1)] The ranker uses two different error functions, RMS for learning and Kendall $\tau$ for stopping criteria. Kendall $\tau$ is not used for learning because it is not continuous or differentiable. Both functions are not consistent as stopping criteria measure the relative ranking, and \textit{RMS} does not, which may lead to incorrect stopping criteria. Enhancing the \textit{RMS} may not also increase the error performance, as illustrated in Fig.~\ref{tab:taurms} in a comparison between the ranker network. evaluation using $\rho$ and \textit{RMS}.
\item[2)] The convergence performance takes many iterations to reach the ranking $\tau\simeq1$ based on the shape of \textit{sigmoid} or \textit{Relu} functions and learning rate as shown in the experiment video link~\cite{videofile} due to the slope shape between -1 or 0 and 1. The prediction probability almost equals the values from -1 or 0 to 1.
\end{itemize}

\subsection{Problem Formulation}
For multi-class and multi-label problems, learning the data's preference relation predicts the class classification and label ranking. i.e. data instance $\Db \in \{x_{1}, x_{2}, \dots , x_{n}\}$. the output labels are predicted as ranked set labels that have preference relations
$\Lb=\{{\lambda _{y}}_{1},\dots,{\lambda _{y}}_{n}\}$.
\textit{PNN} creates a model that learns from an input set of ranked data to predict a set of new ranked data.
The next section presents the initial experiment to rank labels using the usual network structure.

\subsection{Activation Functions}
The usual \textit{ANN} activation functions have a binary output or range of values based on a threshold. However, these functions do not produce multiple deterministic values on the $y$-axis. This paper proposes new functions to slow the differential rate around ranking values on the $y$-axis to solve ranking instability. The proposed functions are designed to be non-linear, monotonic, continuous, and differentiable using a polynomial of the \textit{tanh} function. The step width maintains the stability of the ranking during the forward and backward processes. Moraga~\cite{moraga} introduced a similar multi-valued function. However, the proposed exponential derivative was not applied to an \textit{ANN} implementation. Moraga exponential function is geometrically similar to the step function~\cite{sss}. However, The newly proposed functions consist of \textit{tanh} polynomial instead of exponential due to the difficulty in implementation. The new functions detect consecutive integer values, and the transition from low to high rank (or vice versa) is fast and does not interfere with threshold detection.

\subsubsection{Positive Smooth Staircase (\textit{PSS})}
As a non-linear and monotonic activation function, a positive smooth staircase (PSS) is represented as a bounded smooth staircase function starting from \textit{x}=0 to $\infty$. Thus, it is not geometrically symmetrical around the $y$-axis as shown in Fig. 1. \textit{PSS} is a polynomial of multiple \textit{tanh} functions and is therefore differentiable and continuous. The function squashes the output neurons values during the \textit{FF} into finite multiple integer values. These values represent the preference values from \{\textit{0} to \textit{n}\} where 0 represents the incomparable relation $\bot$ and values from 1 to \textit{n} represent the label ranking. The activation function is given in Eq.~\ref{pss}. \textit{PSS} is scaled by increasing the step width $w$

\begin{figure}[ht]
\begin{center}
\subfloat[] {
\includegraphics[scale=0.7]{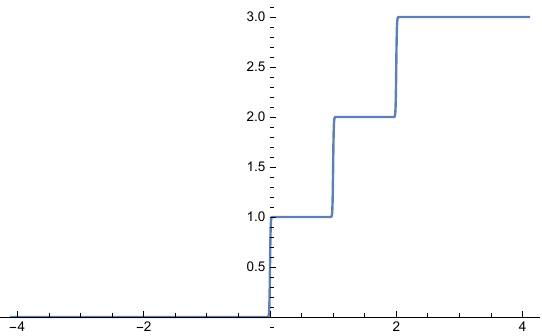}
}
\qquad
\subfloat[] {
\includegraphics[scale=0.68]{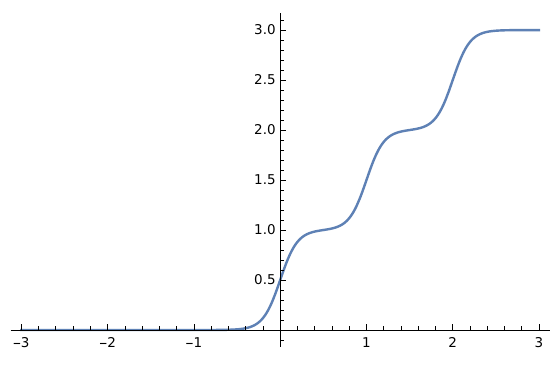}
}

\caption{ $\textit{PSS}$ activation function where $n=3$ and step width $w=1$ and $c=100$ and $5$ in (a) and (b) respectively}
\end{center}
\end{figure}

\begin{equation}\label{pss}
\begin{split}
y=-\frac{1}{2s} \Bigg( \sum_{i=0}^{n-1} \tanh & \left( c(wi-x) \right) -n \Bigg) 
\end{split}
\end{equation}

Where $n$ is the number of stair steps equal to the number of labels to rank, $\textit{w}$ is the step width, and $c$ is the stair curvature $c=100$ and $5$ for the sharp and smooth step, respectively. and $s$ is the scaling factor for reducing the height of each step to range to rank value with decimal place for the regression problems. $s$=10 and $s$=100 for 1 and 2 decimal places, respectively, $s$ is calculated as in Eq.~\ref{maxrange}.

\begin{equation}\label{maxrange}
\begin{split}
n = Y_{max} s
\end{split}
\end{equation}

and $w$ is the step width as shown in Eq.~\ref{stepwidth}.
\begin{equation}\label{stepwidth}
\begin{split}
2b = w(n-1)
\end{split}
\end{equation}

\subsubsection{Smooth Staircase (\textit{SS})}
 The proposed (\textit{SS}) represents a staircase similar to (\textit{PSS}). However, \textit{SS} has a variable boundary value used as a hyperparameter in the learning process. The derivative of the activation function is discussed in section \Romannum{3} and the performance comparison between \textit{SS} and \textit{PSS} is mentioned in Section \Romannum{5}.
 \begin{figure}[ht]
 \begin{center}
\qquad
\subfloat[] {
\includegraphics[scale=0.68]{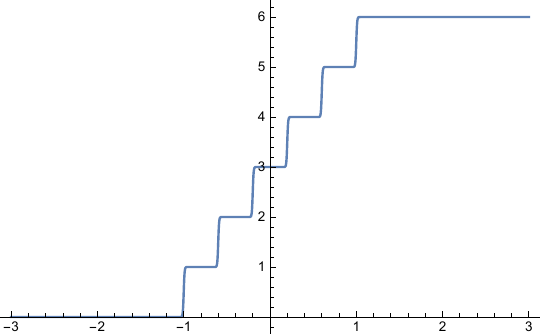}
}
\qquad
\subfloat[] {
\includegraphics[scale=0.68]{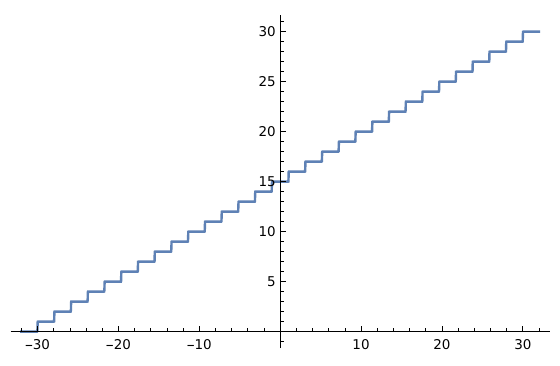}
}
\qquad
\subfloat[] {
\includegraphics[scale=0.68]{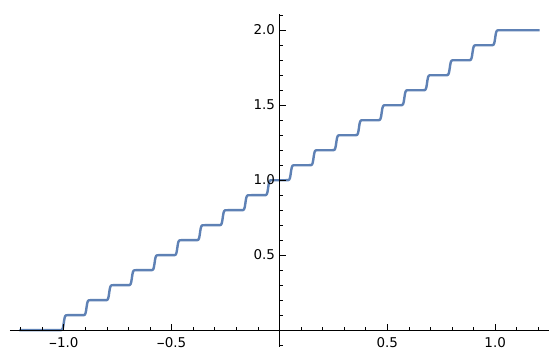}
}
\caption{ $\textit{SS}$ activation function where $n=6$, $30$ and $20$ and boundary $b=1$, $30$ and $1$ and scale factor for the decimal place is $s=1, 1$ and $10$ for ranking/classification, extreme label ranking/classification and regression in (a), (b) and (c) respectively.}
\label{active2}
\end{center}
\end{figure}
 
 The activation function is given in Eq.~\ref{ss}.
\begin{equation}\label{ss}
\begin{split}
y=-\frac{1}{2s} \Bigg( \sum_{i=0}^{n-1} \tanh & \left( c(b-x-wi) \right) -n \Bigg)
\end{split}
\end{equation}
where $c$ is step curvature, $n=$ number of ranked labels, $b$ is the boundary value on the x-axis, and (\textit{SS}) lies between $-b$ and $b$.

where $Y_{max}$ is the max. value to rank. i.e. $Y_{max}$=3 and values have one decimal place. $n$ =30
The (\textit{SS}) function has the shape of smooth stair steps, where each step represents an integer number of label ranking on the $y$-axis from \textit{0} to $\infty$ as shown in Fig. 1, The \textit{SS} step is not flat, but it has a differential slope. The function boundary value on the $x$-axis is from -$b$ to $b$ Therefore, input values must be scaled from -$b$ to $b$. The step width is 1 when n$\simeq2b$. The convergence rate is based on the step width. However, it may take less time to converge based on network hyper parameters. Fig.~\ref{active2} (a) and (b). The \textit{SS} is scaled by increasing the boundary value $b$ 

\subsection{Ranking Loss Function}
Two main error functions have been used for label ranking; Kendall $\tau$~\cite{kendall} and \textit{spearman} $\rho$~\cite{spears}. However, the Kendall $\tau$ function lacks continuity and differentiability. Therefore, the \textit{spearman} $\rho$ correlation coefficient is used to measure the ranking between output labels. \textit{spearman} $\rho$ error derivative is used as a gradient ascent process for \textit{BP}, and correlation is used as a ranking evaluation function for convergence stopping criteria. $\tau_{Avg}$ is the average $\tau$ per label divided by the number of instances $m$, as shown in line 8 of Algorithm 1. \textit{spearman} $\rho$ measures the relative ranking correlation between actual and expected values instead of using the absolute difference of root means square error (\textit{RMS}) because gradient descent of \textit{RMS} may not reduce the ranking error. For example, $\pi_{1}=(1, 2.1, 2.2)$ and $\pi_{2}=(1, 2.2, 2.1)$, have a low \textit{RMS} $=0.081$ but a low ranking correlation $\rho=0.5$ and $\tau=0.3$.

\begin{figure}[ht]
\begin{center}
\pgfplotsset{footnotesize}
\subfloat[Spearman $\rho$] {
\begin{tikzpicture}[scale=0.97]
	\begin{axis}[xlabel=\#iterations,
        grid=major,
        legend style={at={(0.5,0.2)},anchor=west}]
	\addplot coordinates { (0,0.20936963044151782) (10,0.49210465) (20,0.54322) (30,0.54962068) 
    (40,0.57268)(50,0.57380)(60,0.590470)(70,0.594935)
    (80,0.6218696)(90,0.6282692)};

	\addplot coordinates { (0,0.17105) (10,0.54300) (20,0.56271) (30,0.59188) 
    (40,0.590470)(50,0.616885)(60,0.62105)(70,0.62826)
    (80,0.63161)(90,0.651036)};

	\end{axis}

\end{tikzpicture}
}
\subfloat[RMS] {
\begin{tikzpicture}[scale=0.97]
	\begin{axis}[xlabel=\#iterations,
        grid=major,
        legend style={at={(0.5,0.8)},anchor=west}]
	\addplot coordinates { (0,0.38) (10,0.26525) (20,0.28169) (30,0.25821) 
    (40,0.2589)(50,0.24443)(60,0.2553)(70,0.247867)
    (80,0.243491)(90,0.22880)};
    \addlegendentry{\textit{ANN}}
	\addplot coordinates { (0,0.41279) (10,0.29631) (20,0.2594) (30,0.2484) 
    (40,0.24124)(50,0.2396)(60,0.22955)(70,0.209715)
    (80,0.2275)(90,0.24508)};
\addlegendentry{\textit{PNN}}

	\end{axis}
\end{tikzpicture}
}
\caption{Ranker network and \textit{PNN} evaluation in terms of \textit{RMS} and \textit{spearman} correlation error functions}
\label{tab:taurms}
	\end{center}
\end{figure}

Fig ~\ref{tab:taurms} shows the comparison between the initial ranker network and \textit{PNN}; the ranker network uses Kendall $\tau$ which has lower performance as a stopping criterion compared to \textit{PNN} \textit{spearman} because the stopping criteria are based on the \textit{RMS} per iteration; however, \textit{PNN} uses \textit{spearman} for both ranking step and stopping criteria.

The \textit{spearman} error function is represented by Eq.\ref{spearequation}
\begin{equation}\label{spearequation}
\rho=1-\frac{6 \sum_{i=1}^{m}{(y_{i}-yt_{i})^2}}{m(m^2-1)}
\end{equation}
where $y_{i}$, $yt_{i}$, $i$ and $m$ represent rank output value, expected rank value, label index and number of instances, respectively.

\subsection{\textit{PNN} Structure}
\subsubsection{One middle layer}
\begin{figure}[ht]
\SetCoordinates[xAngle =-50]
\SetVertexStyle[MinSize = 8mm]
\SetLayerDistance{-3}
\begin{center}
\begin{tikzpicture}
\node at (0,0) {
\includegraphics[scale=0.6]{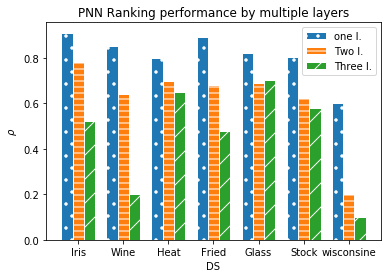}
};
\end{tikzpicture}

\caption{Multiple layer label ranking comparison of benchmark data sets~\cite{Cheng} results using the \textit{PNN} and \textit{SS} functions after 100 epochs and learning rate = 0.007.}
\label{fig:hlayers}
\end{center}
\end{figure}

The \textit{ANN} has multiple hidden layers. However, we propose \textit{PNN} with a single middle layer instead of multi-hidden layers because ranking performance is not enhanced by increasing the number of hidden layers due to fixed multi-valued neuron output, as shown in Fig.~\ref{fig:hlayers}; Seven benchmark data sets ~\cite{Cheng} was experimented using \textit{SS} function using one, two, and three hidden layers with the following hyper parameters; learning rate (l.r.)=0.05, and each layer has neuron $i=100$ and $b=10$). We found that by increasing the number of hidden layers, the ranking performance decreases, and more iterations are required to reach $\rho\simeq1$.
The low performance because of the shape of \textit{SS} produces multiple deterministic values, which decrease the arbitrarily complex decision regions and degrees of freedom per extra hidden layer.

\begin{figure}
\begin{center}
\begin{tikzpicture}[scale=1]
\node at (0,1.5) {\bf \small Preference Neuron};
\node at (0,-1.5) { \footnotesize $\lambda{n}=\varphi_{4}\Big(\sum_{i=1}^{k} a_i.w_i\Big)$};
\draw [thick] (0,0) ellipse (2.3cm and 1cm);
\node at (-2.7,0.1) {\tiny Input };
\node at (-3,-0.2) { \tiny Weights};
\node at (-3.9,1.45) {\tiny{$a_1$}};
\node at (-3,1) { \tiny{$w_1$}};
\node at (-3.9,1) {\tiny .};
\node at (-3.9,0.6) {\tiny .};
\node at (-3.9,0) {\tiny{$a_i$}};
\node at (-3.2,0.2) { \tiny{$w_i$}};
\node at (-3.9,-1) {\tiny .};
\node at (-3.9,-0.6) {\tiny .};
\node at (-3.9,-1.45) {\tiny{$a_k$}};
\node at (-3,-0.9) { \tiny{$w_k$}};
\node at (3.4,1.7) {\tiny{$\lambda{a}$}};
\node at (3.4,0.7) {\tiny{$\lambda{b}$}};
\node at (3.4,-0.3) {\tiny{$\lambda{c}$}};
\node at (3.4,-1.2) {\tiny{$\lambda{d}$}};
\node at (2,0) {\tiny{$a_j$}};
\node at (-1.4,0) {\tiny {$\sum_{i=1}^{k} a_i.w_i$} };
\node at (0,0) {\tiny{$\varphi$}};
\node at (0.55,0) {
\includegraphics[scale=0.2]{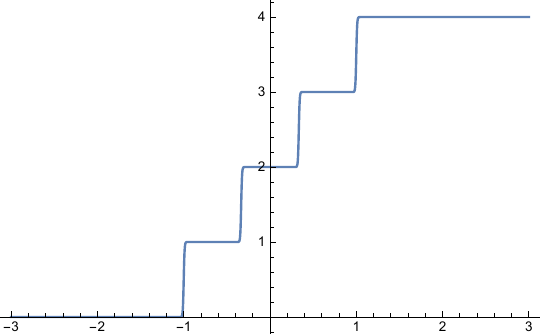}
};
\draw [thick] (1.7,0) -- (1.7,-0.68);
\draw [thick] (1.7,0) -- (1.7,0.68);
\draw [thick] (-0.6,0) -- (-0.6,-0.97);
\draw [thick] (-0.6,0) -- (-0.6,0.97);
\draw [->,>=stealth] (1.9,0.57) -- (3.7,1.6);
\draw [->,>=stealth] (2.27,0.2) -- (3.7,0.6);
\draw [->,>=stealth] (2.27,-0.2) -- (3.7,-0.6);
\draw [->,>=stealth] (1.9,-0.57) -- (3.7,-1.6);
\draw [->,>=stealth] (-3.7,1.6) -- (-2,0.5);
\draw [->,>=stealth] (-3.7,0) -- (-2.3,0);
\draw [->,>=stealth] (-3.7,-1.6) -- (-2,-0.5);
\end{tikzpicture}
\caption{The structure of preference neuron where $\varphi_{n=4}$.}
\label{fig:pnn}
\end{center}
\end{figure}

\subsubsection{Preference Neuron} Preference Neuron are a multi-valued neurons uses a \textit{PSS} or \textit{SS} as an activation function. Each function has a single output; however, \textit{PN} output is graphically drawn by $n$ number of arrow links that represent the multi-deterministic values. The \textit{PN} in the middle layer connects to only $n$ output neurons $stp=n+1$; where $stp$ is the number of \textit{SS} steps. The \textit{PN} in the output layer represents the preference value. The middle and output \textit{PN}s produce a preference value from 0 to $\infty$ as illustrated in Fig.~\ref{fig:pnn}.

The \textit{PNN} is fully connected to multiple-valued neurons and a single-hidden layer \textit{ANN}. The input layer represents the number of features per data instance. The hidden neurons are equal to or greater than the number of output neurons, $H_{n}\geq\Lb_{n}$, to reach error convergence after a finite number of iterations. The output layer represents the label indexes as neurons, where the labels are displayed in a fixed order, as shown in Fig.~\ref{fig:typea-1}.

\SetCoordinates[xAngle =-35]
\SetVertexStyle[MinSize = 8mm]
\SetLayerDistance{-3}
\SetPlaneWidth{5.5}
\SetPlaneHeight{5.5}
\begin{figure}[ht]
\begin{center}

\begin{tikzpicture}[scale=1.1,multilayer=2d,rotate=-90]
\begin{Layer}[layer=2]
\node[draw] at (1.6,-3.5) {\footnotesize One Instance};

 \node at (2.6,-0.7)[rotate=90] {\input{TypeA.tikz}};
\node at (-2.2,1)[rotate=90]{\footnotesize Output layer};
\node at (-2.2,-1)[rotate=90]{\footnotesize Middle layer};
\node at (-2.2,-3)[rotate=90]{\footnotesize Input layer};

\node at ( -1,-2.5)[rotate=90]{\footnotesize \#16n.};
\node at ( -1.9,-1)[rotate=90]{\footnotesize \#300h.n.};
\node at ( -1,1)[rotate=90]{\footnotesize \#16n.};

\node at (0.6,2) [draw]  {\footnotesize 3} ;
\node at (1.6,2) [draw]  {\footnotesize 1};
\node at (2.5,2) [draw]  {\footnotesize 4};
\node at (3.5,2) [draw]  {\footnotesize 5};
\node at (4.2,2) [draw]  {\footnotesize 2};

\end{Layer}
\end{tikzpicture}
\caption{\textit{PNN} where $\varphi_{n=16}$, $f_{in}=16$ and $\lambda_{out}=16$, per $\langle x_{1} , \pi_{1}\rangle$, $\Lb\in\{\lambda_{a},\lambda_{b},\lambda_{c},\lambda_{d}$\} where $\pi_{1}=\{1,2,3,4,\dots, 16\}$.}
\label{fig:typea-1}
\end{center}
\end{figure}

 The \textit{ANN} is scaled up by increasing the hidden layers and neurons; however, increasing the hidden layers in \textit{PNN} does not enhance the ranking correlation because it does not arbitrarily increase complex decision regions and degrees of freedom to solve more complex ranking problems. This limitation is due to the multi-semi discrete-valued activation function, limiting the output data variation. Therefore, instead of increasing the hidden layer, \textit{PNN} is scaling up by increasing the number of neurons in the middle layer and scaling input data boundary value and increasing the \textit{PSS} step width and \textit{SS} boundaries which are equal to the input data scaling value, which leads to increased data separability.

\textit{PNN} reaches ranking $\rho\simeq1$ after 24 epochs compared to the initial ranker network that reaches the same result in 200 iterations, The video link demonstrates the ranking convergence as shown in Fig.~\ref{ranker} and video~\cite{videofile}. A summary of the three networks is presented in Table \Romannum{1}.

The output labels represent the ranking values. The differential \textit{PSS} and \textit{SS} functions to accelerate the convergence after a few iterations due to the staircase shape, which achieves stability in learning.
\textit{PNN} simplifies the calculation of \textit{FF} and \textit{BP}, and updates weights into two steps due to single middle layer architecture. Therefore, the batch weight updating technique is not used in \textit{PNN}, and pattern update is used in one step. The network bias is low due to the limited preference neuron output of data variance; thus it is not calculated. Each neuron uses the \textit{SS} or \textit{PS} activation function in \textit{FF} step, and calculates the preference number from 1 to \textit{n}, where \textit{n} is the number of label classes. During \textit{BP}. The processes of \textit{FF} and \textit{BP} are executed in two steps until $\rho_{Avg}\simeq1$ or the number of iterations reaches ($10^6$) as mentioned in the algorithm section.

The \textit{SS} step width decreases by increasing the number of labels; thus, we increase function boundary $b$ to increase the step width to $\simeq1$ to make the ranking convergence; In addition, a few complex data sets may need more data separability to enhance the ranking. Therefore, we use the $b$ value as a hyperparameter to keep the stair width $>=1$ and normalize input data from $-b$ to $b$.

\begin{figure}[ht]
\begin{center}
\pgfplotsset{footnotesize}
\begin{tikzpicture}[scale=1.2]

\node at (-1,1)[] {\textit{$\tau$}};
	\begin{axis}[title={\footnotesize{Ranking Convergence}},xlabel=\#iterations,grid=major,
        legend style={at={(1.1,0.4)}}]
	\addplot coordinates { (0,0)(60,0)(61,0.62)(145,0.62) (150,1)(200,1)};
    \addlegendentry{\textit{Ranker NN}}
	\addplot coordinates { (0,0)(25,0) (35,0.8)(50,1)(200,1)};
\addlegendentry{\textit{PNN}}
	\end{axis}

\end{tikzpicture}
\caption{The structure used in both ranker \textit{ANN} and \textit{PNN} where $\varphi_{n=3}$, $f_{in}=3$ and $\lambda_{out}=3$, per $\langle x_{1} , \pi_{1}\rangle$, $\Lb\in\{\lambda_{a}, \lambda_{b}, \lambda_{c}$\} where $\pi_{1}=\{1, 2, 3\}$. and comparison of the convergence for both NN's. The demo video of convergence of two NN in the link~\cite{videofile}.}
\label{ranker}
\end{center}
\end{figure}
Table \Romannum{1} shows a brief comparison between Ranker \textit{ANN} and \textit{PNN}.
\begin {table}[ht]
\begin{center}
\label{tabcomparison}
\caption {\textit{ANN} types used in initial experiment.}
\label{tab:networktypes}
\begin{tabular}{|c|c|c|c|c|c|}\hline
\textbf{Type} &\textbf{Ranker \textit{ANN}}& \textbf{\textit{PNN} }\\ \hline

\textbf{Activation Fun.}&\textit{ReLU,Sigmoid}&PSS, SS\\ \hline
\textbf{Gradient}&Descent&Ascent\\ \hline
\textbf{Objective Fun.}& \textit{RMS}&$\rho$\\ \hline
\textbf{Stopping Criteria.}&$\tau$&$\rho$\\ \hline

\end{tabular}
\end{center}
\end{table}

The following section describes the data preprocessing steps, feature selections, and components of \textit{PN}.

\section{\textit{PN} Components}
\subsection{Image Preprocessing}
\subsubsection{Greyscale Conversion}
 Data scaling as red, green, and blue (\textit{RGB}) colors is not considered for ranking because \textit{PN} measures the preference values between pixels. Thus, The image is converted from \textit{RGB} color to Greyscale.
\subsubsection{Pixels' Sorting}
Ranking the image from $\pi=\{\lambda_{1},..,\lambda_{m}$\} to $\pi=\{\lambda_{1},..,\lambda_{k}$\} where the maximum greyscale value 
$\lambda_{m}=255$ and $\lambda_{k}$ is the maximum ranked pixel value
as illustrated in Fig.~\ref{pixelranking1} (a).
\begin{figure}[ht]
\begin{tikzpicture}[]
\node at (-2,0)[scale=0.8] {\input{grid1.tikz}};
\node at (2,0)[scale=0.8] {\input{grid2.tikz}};
\node at (1.4,-3)[scale=0.8] {\input{grid3.tikz}};
    
\node at ( -1.9,1.9)[]{\footnotesize 28 X 28};
\node at ( -1.9,1.5)[]{\footnotesize 0-255};

\node at ( 2,1.9)[]{\footnotesize 28 X 28};
\node at ( 2,1.5)[]{\footnotesize 1-156};

\node at ( 1.7,-1.5)[]{\footnotesize 3 X 3};
\node at ( 1.7,-1.8)[]{\footnotesize 1-23};
  
\node at ( 2.6,-2.7)[]{\footnotesize 3 X 3};
\node at (4.4,0.4)[]{\small (a)}; 
\node at (4.4,-2.7)[]{\small (b)}; 

\node at ( 2.6,-3)[]{\footnotesize 1-9};

  \node at (-0.5,0) (r1) {};
  \node  at (0.6,0) (r2) {};
  
  \node at (1,-1.5) (r3) {};
  \node  at (1,-2.2) (r4) {};
    \draw[->,>=stealth ]  (r1) edge (r2); 
   \draw[->,>=stealth ]  (r3) edge (r4); 
\end{tikzpicture}
\caption{Image pixel sorting for the flattened windows in (a) and (b) respectively.}
\label{pixelranking1}
\end{figure}

\subsubsection{Pixels Averaging}
\begin{figure}[ht]
\begin{center}
\subfloat[$\rho=0.216$] {
\includegraphics[scale=0.5]{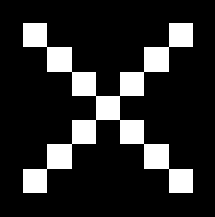}
\includegraphics[scale=0.5]{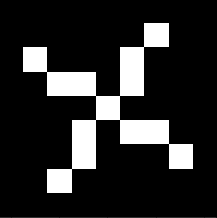}
}
\subfloat[$\rho=0.79$] {
\includegraphics[scale=0.31]{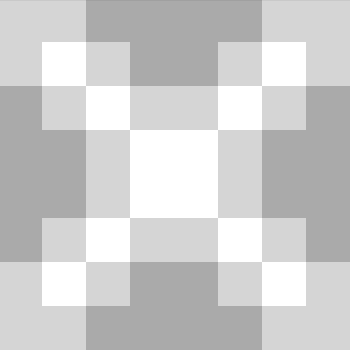}
\includegraphics[scale=0.31]{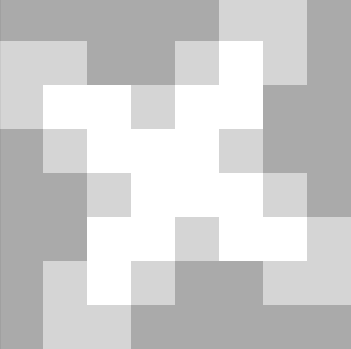}
}
\qquad
\subfloat[$\rho=0.137$] {
\includegraphics[scale=0.28]{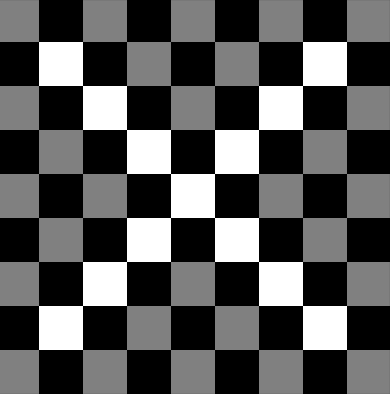}
\includegraphics[scale=0.28]{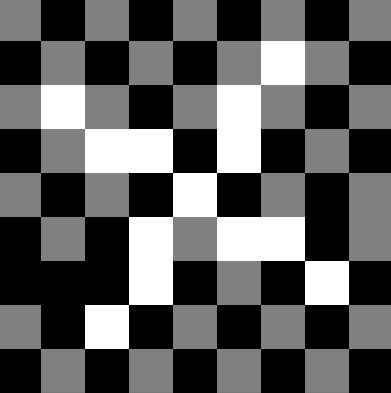}
}
\subfloat[$\rho=0.75$] {
\includegraphics[scale=0.32]{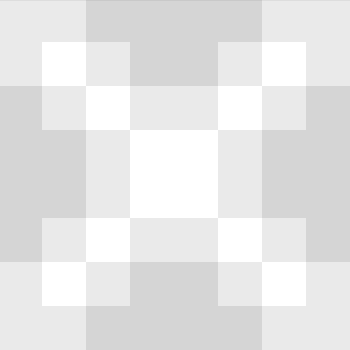}
\includegraphics[scale=0.32]{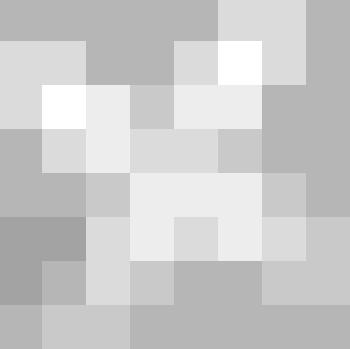}
}
\qquad
\subfloat[$\rho=-0.18$] {
\includegraphics[scale=0.51]{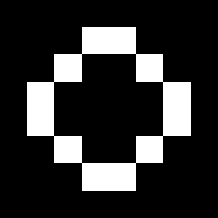}
\includegraphics[scale=0.51]{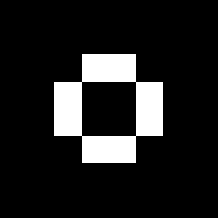}
}
\subfloat[$\rho=0.71$] {
\includegraphics[scale=0.58]{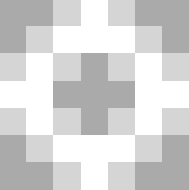}
\includegraphics[scale=0.6]{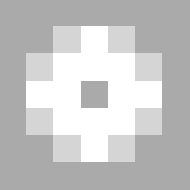}
}
\caption{Sample of moving objects in (a) and (b) without and with averaging by window 2x2. The ranking of two flattened images are $\rho=0.216$ and $0.79$ in (a) and (b), respectively. Sample of moving noisy object in (c) and (d) without and with image averaging by a window of 2x2. The ranking of two flattened images are $\rho=0.137$, $0.75$ and $0.75$ in (c) and (d) respectively. ranking scaled circle in (e) and (f), respectively.}
\label{fig:imgavrg}
\end{center}
\end{figure}

Ranking image pixels has an almost low ranking correlation due to noise, scaling, light, and object movement; therefore, window averaging is proposed by calculating the mean of pixel values of the small flattened window size of 2x2 of 4 pixels as shown in Fig.~\ref{fig:imgavrg}. The overall image $\rho$ of pixels increased from 0.2 to 0.79 in (a and b), from 0.137 to 0.75 for noisy images in (s and d), and scaled images from -0.18 to 0.71 in (e and f).

\begin{figure}[ht]
\begin{center}
\begin{tikzpicture}
\node at (-1,-4){\includegraphics[scale=1.1]{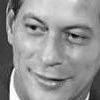}};
\node at (1.1,-4){\includegraphics[scale=1.1]{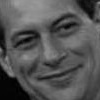}};
\node at (-5.7,-4){\includegraphics[scale=.28]{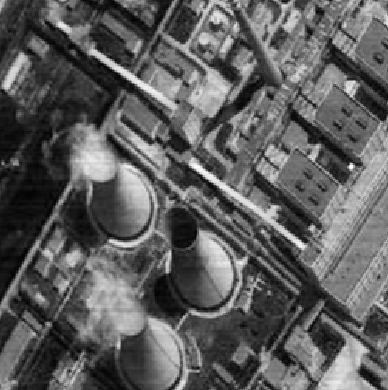}};
\node at (-3.6,-4){\includegraphics[scale=.28]{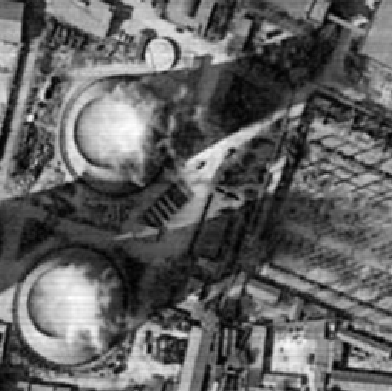}};
\node at (-0.5,-3.6) [draw,green] {};
\node at (1.5,-3.5) [draw,green] {};
\node at (-1.3,-3.6) [draw,blue] {};
\node at (0.8,-3.7) [draw,blue] {};
\node at (-5.7,-4.4) [shape=rectangle,draw=red,minimum width=20pt, minimum height=20pt]  {};
\node at (-3.9,-3.7) [shape=rectangle,draw=red,minimum width=20pt, minimum height=20pt]  {};

\node [draw ,green] at (1.3,-2.5){\includegraphics[scale=.05]{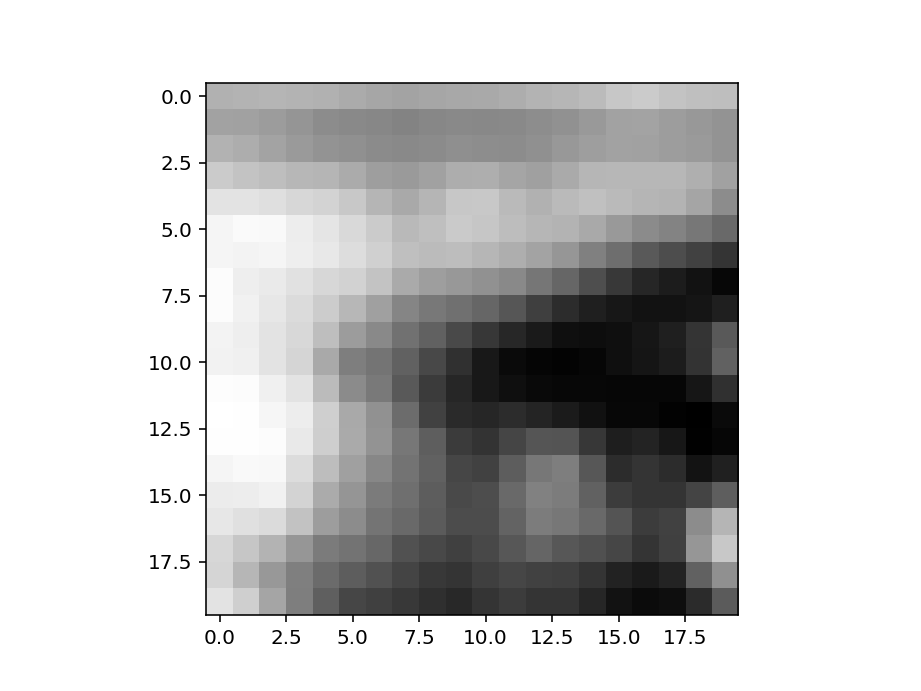}};
\node [draw ,green] at (0,-2.5){\includegraphics[scale=.05]{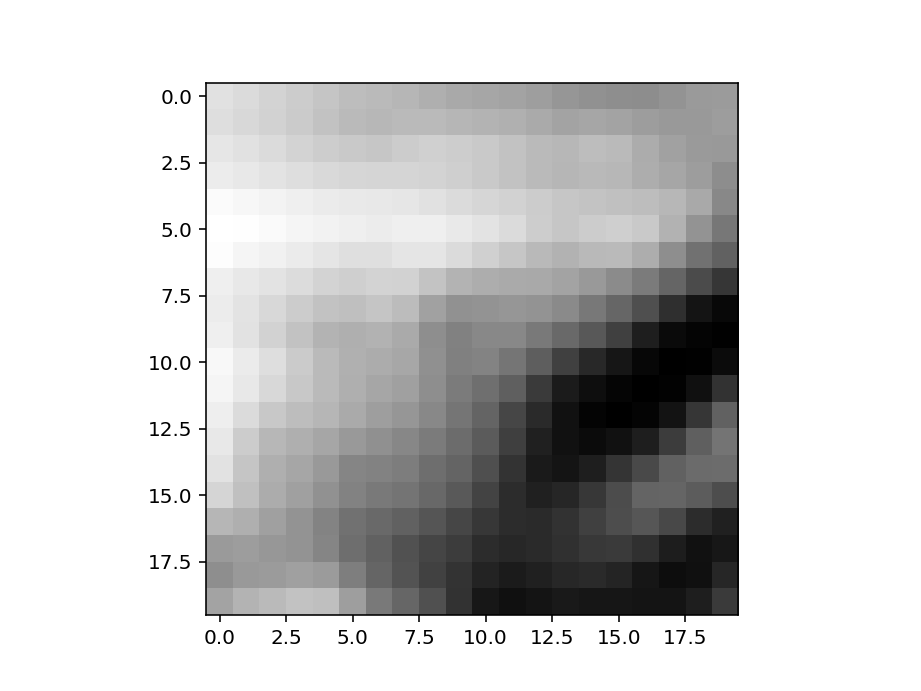}};

\node[draw ,blue] at (-3,-2.5){\includegraphics[scale=.05]{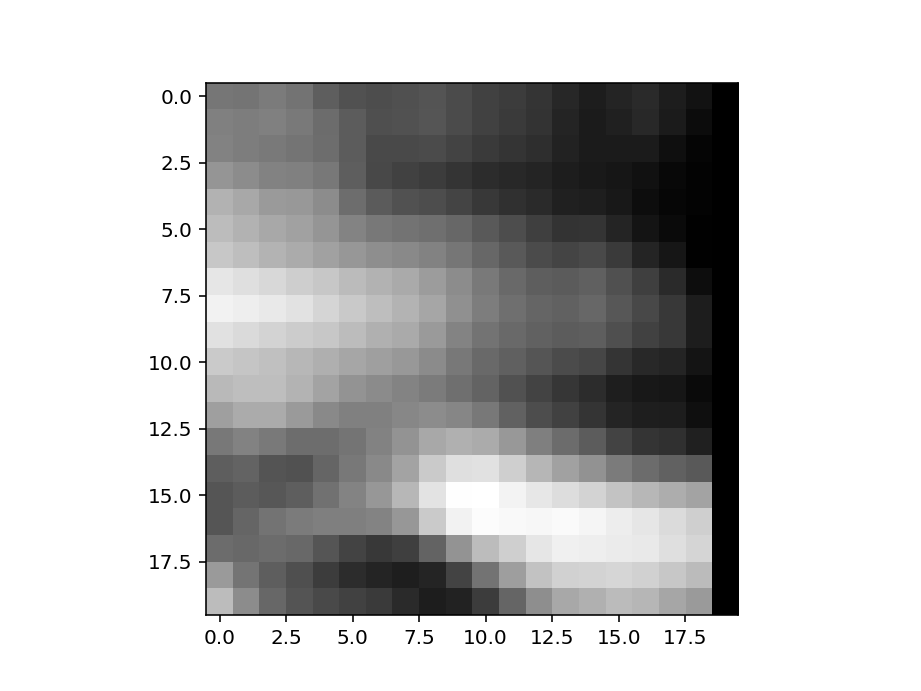}};
\node[draw ,blue] at (-1.6,-2.5){\includegraphics[scale=.05]{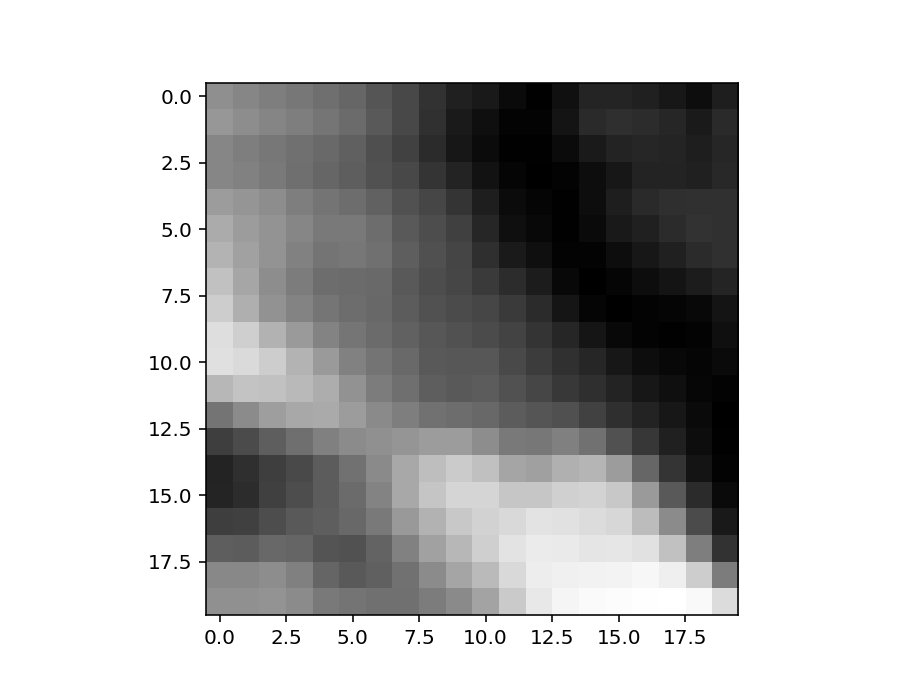}};

\node[draw ,red] at (-5.8,-2.5){\includegraphics[scale=.4]{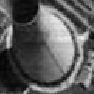}};
\node[draw ,red] at (-4.7,-2.5){\includegraphics[scale=.3]{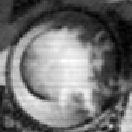}};

\node at (0.5,-1.8) [] {\footnotesize  $\rho=0.84$ };
\node at (-2.7,-1.8) [] {\footnotesize  $\rho=0.81$ };
\node at (-5,-1.8) [] {\footnotesize  $\rho=0.6$ };
\end{tikzpicture}
\caption{Detecting the similarity in remote sensing and face recognition by ranking the image pixels after averaging the pixels using a 2x2 window.}
\label{pixelranking}
 \end{center}
 \end{figure}

The two approaches, Pixel ranking and Averaging has been tested in remote sensing and faces images to detect the similarity, and it shows high ranking correlations using different window size as shown in Fig~\ref{pixelranking}. It detects the high correlation by starting from the large window size $=$ image size. It reduces the size and scans until it reaches the highest correlation.

\subsection{Feature Selection By Attention}
Feature selection for the kernel proceeded by selecting the features with a high group of pixel ranking variations indicating the importance of the scanned kernel area. This kind of hard attention makes the selection based on the threshold of pixel ranking values. to reduce the dimension of the input image. 
\subsection{Feature Extraction}
 This paper proposes a new approach for image feature selection based on the preference values between pixels instead of the convolution of pixels array as implemented in \textit{CNN}. The \textit{PN}'s features are based on ranking computational space. Therefore, the kernel window size is considered a factor for feature selection.
 
\subsubsection{Pixels Resorting}
The flattened window's values are sorted for each kernel window in the image. The Fig.~\ref{pixelranking1} (b) shows the window size 3X3 range from $\lambda_{k_{1}}=23$ to $\lambda_{k_{2}}=9$. Pixel sorting reduces the data margin, Thus, it reduces the computational complexity.

\subsubsection{Weighted Ranker Kernel}
The kernel weights are randomly initialized from -0.05 to 0.05. The kernel learns the features by BP of its weights to select the best feature. the partial change in the kernel is calculated by differentiating the \textit{spearman} correlation as in Eq.~\ref{s7}
\begin{equation}\label{s7}
dKw=2\cdot Img_{w}-d\rho\cdot \frac{n^3-n}{-6}
\end{equation}
Different kernel sizes could be used for big images' size. We use three different kernels to capture the relations between different features.
\subsubsection{Max Pooling}
Max. pooling is used to reduce the features map's size and select the highest correlation values to feed to the \textit{PNN}.

\SetCoordinates[xAngle =-35]
\SetVertexStyle[MinSize = 8mm]
\SetLayerDistance{-3}
\SetPlaneWidth{5.5}
\SetPlaneHeight{5.5}
\begin{figure*}[ht]
\begin{center}
\begin{tikzpicture}[scale=0.86,multilayer=2d,rotate=-90,
circle/.style = {shape=circle, aspect=2.2, draw} ]
\begin{Layer}[layer=2]
\node at (-5,-9)[rotate=90] {\input{flowchart.tikz}};
\node at (2.6,-3.5)[rotate=90] {\input{TypeB.tikz}};
    
\node at (-1.8,-18.6){\includegraphics[angle=90,scale=1]{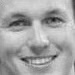}};
 \node at (2.9,-19){\includegraphics[angle=90,scale=1.2]{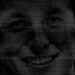}};
    
\node at (-1,-14.7)[] {\input{Kernel3.tikz}};
\node at (3,-13.5)[rotate=-90] {\input{Kernel2.tikz}};
\node at (7,-14.5)[] {\input{Kernel1.tikz}};
    
\node at (-1.2,-9.3)[rotate=90] {\input{spearmangrid1.tikz}};
\node at (2.5,-9.3)[rotate=90] {\input{spearmangrid2.tikz}};
\node at (7,-9.8)[rotate=90] {\input{spearmangrid3.tikz}};
  
\node at (-3.4,0.3)[rotate=90]{\textbf{Output l.}};
\node at (-3.4,-1.7)[rotate=90]{\textbf {Middle l.}};
\node at (-3.4,-3.5)[rotate=90]{\textbf {Input l.}};
\node at ( -3.4,-19)[rotate=90]{\textbf{Image Reprocessing}};
\node at ( -3.4,-15)[rotate=90]{\textbf{Feature Extraction}};
\node at (-3.4,-9.2) [rotate=90] {\textbf {Ranking Layer}};
\node at (-3,-9.2) [rotate=90] { $\rho$ Feature Maps};

\node at (-2.6,-9.2) [rotate=90] {9X9};
\node at (0.5,-9.2) [rotate=90] {19X19};
\node at (4.5,-9.2) [rotate=90] {26X26};

\node at (-3.4,-5.7)[rotate=90]{\textbf {Max. Pooling}};
\node at (-1.5,-6)[rotate=90]{\input{maxpoolinggrid1.tikz}}; 
\node at (2,-6)[rotate=90] {\input{maxpoolinggrid2.tikz}}; 
\node at (6,-6)[rotate=90] {\input{maxpoolinggrid3.tikz}}; 
\node at (-1.2,-19.6) [draw] [circle,rotate=90]  {2};
\node at (-0.5,-19.6) [draw] [circle,rotate=90]  {3};
\node at (0.2,-19.6) [draw] [circle,rotate=90]  {4};
\node at (.9,-19.6) [draw] [circle,rotate=90]  {5};
\node at (5,-19) [draw] [circle,rotate=90]  {6};
\node at (6,-18) [draw] [circle,rotate=90]  {7};
\node at (7,-17) [draw] [circle,rotate=90]  {8};
\node at (8,-16) [draw] [circle,rotate=90]  {9};

\node at (-1,-12.7) [draw] [circle,rotate=90]  {10};
\node at (3.2,-12) [draw] [circle,rotate=90]  {10};
\node at (7.6,-12.7) [draw] [circle,rotate=90]  {10};

\node at (4,-17.9) [shape=rectangle,draw=white,minimum width=25pt, minimum height=25pt]  {};
\node at (2.7,-18.15) [shape=rectangle,draw=white,minimum width=40pt, minimum height=40pt]  {};
\node at (2.4,-18.52) [shape=rectangle,draw=white,minimum width=60pt, minimum height=60pt]  {};

\node at( 5,-16.5)[rotate=90]{\footnotesize $\rho$};
\node[rotate=90]at(3,-15.5)(r3){\footnotesize $\rho$};
\node[rotate=90]at(0.5,-16.4)(r3){\footnotesize $\rho$};

\draw[dotted] (-2.7,-12) -- (8.5,-12);
\draw[dotted] (-2.7,-7.7) -- (8.5,-7.7);
\draw[dotted] (-2.7,-4.8) -- (8.5,-4.8);

\draw[dotted] (-2.7,-2.5) -- (8.5,-2.5);
\draw[dotted] (-2.7,-0.7) -- (8.5,-0.7);

 \node  at (-1,-12) (a) {};
 \node  at (-1,-11) (b) {};

  \node  at (2,-12) (w1) {};
  \node  at (2,-11) (w2) {};

  \node  at (7,-13) (w3) {};
  \node  at (7,-12) (w4) {};
  
  \node  at (0.2,-19) (p1) {};
  \node  at (1.6,-19) (p2) {};

 \draw[->,>=stealth ]  (a) edge (b); 
 \draw[->,>=stealth ]  (w1) edge (w2); 
 \draw[->,>=stealth ]  (w3) edge (w4);
 
  \node  at (3.5,-17.55) (p3) {};
  \node  at (6.4,-13.65) (p4) {};
   \node  at (4.3,-18.4) (p5) {};
   \node  at (7.85,-15.2) (p6) {};
 
 \draw[-]  (p3) edge (p4); 
  \draw[-]  (p5) edge (p6); 
 
   \node  at (2,-17.55) (p7) {};
  \node  at (1.9,-14.5) (p8) {};
   \node  at (3.4,-17.55) (p9) {};
   \node  at (3.9,-14.6) (p10) {};
   
\draw[-]  (p7) edge (p8); 
 \draw[-]  (p9) edge (p10); 
   
  \node  at (1.5,-19.7) (p11) {};
  \node  at (-2.5,-16) (p12) {};
   
  \node  at (3.45,-17.55) (p13) {};
  \node  at (0.05,-13.4) (p14) {};
   
\draw[-]  (p11) edge (p12); 
 \draw[-]  (p13) edge (p14);  

\node at (-2.8,-14.5)[rotate=90]{\footnotesize w.kernel - 20X20};
\node at (1.5,-13.7)[rotate=90]{\footnotesize w.kernel - 10X10};
\node at (6,-14.5)[rotate=90]{\footnotesize w.kernel - 3X3};

  \node  at (6.5,-15) (c1) {};
  \node  at (6.5,-13.5) (d1) {};
  \node  at (7.3,-11.7) (c) {};
  \node  at (7.3,-9.2) (d) {};
  \node at (-0.6,-20.1) (r1) {};
  \node at (0.7,-20.1) (r2) {};
   \draw[->,>=stealth ]  (r1) edge (r2); 

\node at (0.9,0.3) [draw]  {\footnotesize 0} ;
\node at (1.9,0.3) [draw]  {\footnotesize 1};
\node at (2.8,0.3) [draw]  {\footnotesize 0};
\node at (3.8,0.3) [draw]  {\footnotesize 0};
\node at (4.5,0.3) [draw]  {\footnotesize 0};
\end{Layer}
\end{tikzpicture}
\caption{The \textit{PN} structure has three kernels and three \textit{PNN}s where $\varphi_{n=2}$, $f_{1in}=16, f_{2in}=81, f_{3in}=169$ and $\lambda_{out}=15$, per $\langle x_{1} , \pi_{1}\rangle$,$\pi\in\{\lambda_{1},\lambda_{2},\lambda_{3}\cdots,\lambda_{15}$\}.}
\label{fig:typea-2}
\end{center}
\end{figure*}
\subsection{\textit{PN} Structure}
\textit{PN} is the deep learning structure of \textit{PNN} for image classification. It consists of five layers, a ranking features map and a max. pooling and three \textit{PNN} layers. \textit{PN} has one or multiple different sizes of \textit{PNN}s connected by one output layer. Each \textit{PNN} has \textit{SS} or \textit{PSS} where $\varphi_{n=2}$ for binary ranking to map the classification. The number of output neurons is the number of classes. The structure is shown in Fig~\ref{fig:typea-2}. \textit{PN} have one or more ranker kernels with different sizes, Each kernel has one corresponding \textit{PNN}. 
\textit{PN} uses the weighted kernel ranking to scan the image and extract the features map of \textit{spearman} correlation values of the kernel with the scanned ranked image window as
$\rho(\pi_{k},\pi_{w})$ where $\pi_{k}$ is the kernel preference values and $\pi_{w}$ is the scanned window image preference values.
Each kernel scans the image by one step and creates a \textit{spearman} features list. Max. Pooling is used to minimize the feature map used as input to \textit{PNN}.

One 5X5 kernel is used for fashion \textit{Mnist} data set~\cite{mnist}. Three kernels with sizes (3, 10, and 20) are used for \textit{CFAR-100}~\cite{cfar}.

\subsection{Choosing The Kernel Size}

Kernel size is chosen based on the hard attention of the highest group of pixels that has high ranking variation. The process scans the image sequentially starting from a small size to find the size with the highest pixels ranking variation. For example for the Mnist dataset where the image has a size of 28X28, The meaningful features are extracted using kernel sizes 10x10, 15x15, 20x20 and 25x25.

\section{Algorithms}
\subsection{Baseline Algorithm}
Algorithm 1 represents the three functions of the network learning process; feed-forward (\textit{FF}), \textit{BP}, and updating weights (\textit{UW}). Algorithm 2 represents the learning flow of \textit{PN}. Algorithm 3 represents the simplified BP function in two steps.

\begin{algorithm}[ht]
\KwData{$\Db \in \{x_{1}, x_{2}, \dots ,  x_{d} \}$}
\KwResult{ $\pi\in \{{\lambda _{y}}_{1},\dots,{\lambda _{y}}_{n}\}$}
Randomly initialize weights $\omega_{i,j} \in \{-0.05,0.05\}$\\
\Repeat{$\rho_{Avg}$ = 1 or $\#iterations \geq 10^6$}{
\ForAll{$\langle x_{i} , \pi_{i}\rangle \in \Db$ }{
      $a_{i}|_{l-1}=\sum_{i=1}^{m} \varphi\big( {a_{i} \cdot \omega_{i}\big)|_{n}}$ // FF\\
      \textit{PNN} BP()\\
      ${\omega_{i}}_{new}={\omega_{i}}_{old} - \eta \cdot \delta_{i}$ //UW\\
}
}
\caption{\textit{PNN} learning flow}
\end{algorithm}

\begin{algorithm}[ht]
  Converting image to greyscale \\
  Flattening image\\
  Pixels sorting \\ 
  2D Image  \\ 
  Pixels averaging by a 2X2 window \\ 
  Flattening image\\
  Select one/more kernel sizes.\\
  Random init. Kernel $K\omega_{x,y} \in \{-0.05,0.05\}$\\
  Random init. \textit{PNN} $\omega_{i,j} \in \{-0.05,0.05\}$\\
  \Repeat{$\rho_{Avg}$ = 1 or $\#iterations \geq 10^6$}{
  2D Image \\
  Scanned window pixel ranking $Img_{w}$ \\ 
  Compute $\rho(Img_{w},Kw)$ feature map\\
  Max. Pooling.\\
  Flattening image\\
  \textit{PNN} FF()\\
  \textit{PNN} BP()\\
  \textit{PNN} UW()\\
  Max. Pooling BP()\\
  Ranker kernel BP and UW()\\
}
  
\caption{\textit{PN} Learning flow}
\end{algorithm}

\begin{algorithm}[ht]
  Step 1: 
     \For{each $\textit{pn}_{i}$ in Output layer}{ 
       ${Err}_i=\rho=-6\cdot\frac{(2yt_{i}-y_{i})}{m(m^2-1)}$ //\textit{spearman} error \\
        $\delta_{i}={Err}\cdot\varphi\prime $ \\
      }
 Step 2:
 \For{each $\textit{pn}_{i}$ in middle layer}{ 
     ${Err}_i=\sum_{k=0}^{m}{\omega_{k} }\cdot \delta_{k}$\\
        $\delta_{i}={Err}\cdot\varphi\prime $ \\
 }
\caption{\textit{PNN} BP}
\end{algorithm}

\subsection{Ranking Visualization}
\begin{figure}[ht]
\begin{center}
\subfloat[] {
\includegraphics[scale=0.54]{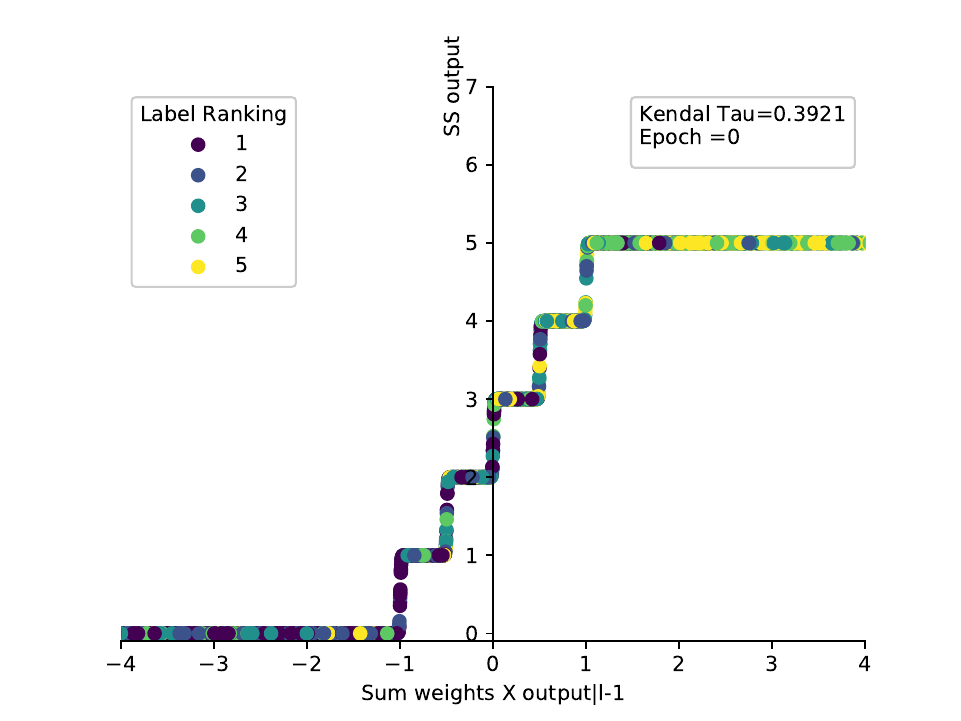}
}
\newline
\subfloat[] {
\includegraphics[scale=0.54]{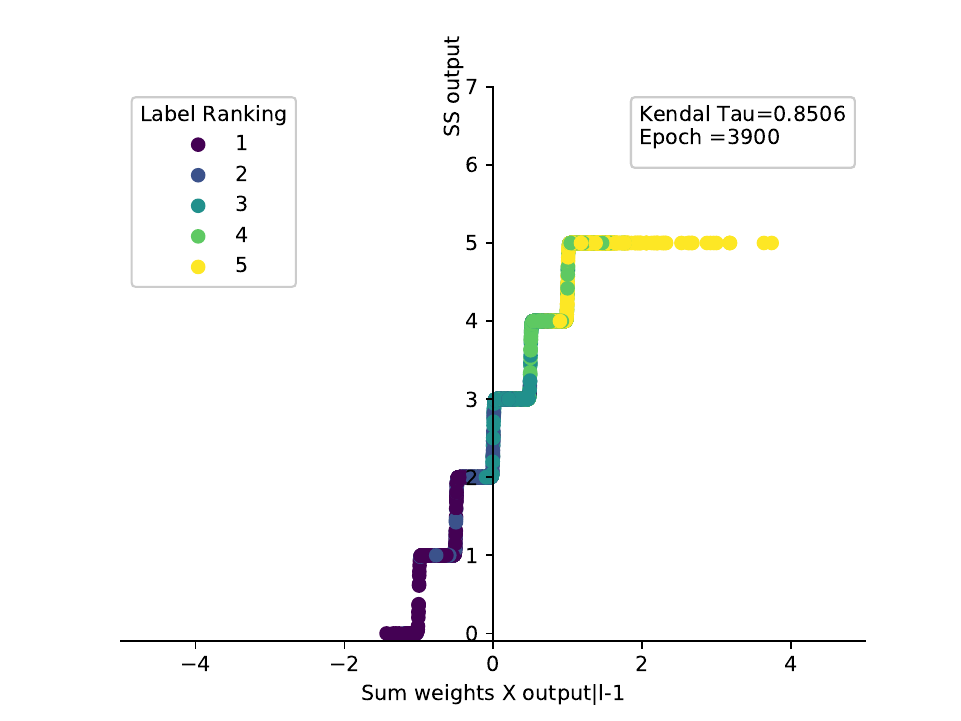}
}
 \caption{Visualizing the ranking of stock dataset~\cite{Cheng} has five labels using \textit{SS} activation function of stock data set at epoch 0 and 3900 in (a) and (b) respectively.
 \label{fig:visconv}}
\end{center}
\end{figure}

\textit{PNN} ranking convergence is visualized using the \textit{SS} function by displaying the normalized input data points with corresponding actual ranked five labels represented in 5 different colours, The plotting of input value and \textit{SS} output values per iteration is shown in Fig.~\ref{fig:visconv}, which illustrates the distribution of \textit{SS} output values against the actual colour values at iterations 0 and 3900 and $\tau$ is enhanced from 0.39 to 0.85.
\subsection{Complexity Analysis}
\subsubsection{Time Complexity}
\begin{itemize}
\item {\textit{FF}} time complexity corresponds to \textit{FF} of middle and output layers, and $m$ and $n$ are the number of nodes in the middle and output layers. $W_m$ and $W_o$ are weighted matrix and $SS_t$ is the activation function of number of instances $t$. The time complexity in Eq.~\ref{ffcom}
\begin{equation}\label{ffcom}
\mathcal{O}(m\cdot o\cdot t) 
\end{equation}
\item \textit{BB} starts with calculating the error of output layer $E_{ot}=\rho_o'$
$Delta_{o}=E_{ot}\cdot SS'$ and 
$Delta_{m}=E_{mt}\cdot SS'$
then UW
\begin{equation}
W_m=W_m-Delta_{m}
\end{equation}
This time complexity is then multiplied by the number of epochs $p$
\begin{equation}
\mathcal{O}(p\cdot m\cdot o\cdot t)
\end{equation}
\end{itemize}
\subsubsection{Input Neurons}
The number of \textit{PN} input neurons is represented by Eq.~\ref{inputno}
\begin{equation}\label{inputno}
\#Input = (Img_{w}-K_{w}+1)\cdot(Img_{h}-K_{h}+1)
\end{equation}
where $w$ and $h$ are width and height of kernel and image.
\section{Network Evaluation}
This section evaluates the \textit{PNN} against different activation functions and architectures. All weights are initialized = 0 to compare activation functions and \textit{A} and \textit{B} have the same initialized random weights to evaluate the structure.

\subsection{Activation Functions Evaluation}
\textit{PNN} is tested on iris and stock data sets using four activation functions. \textit{SS}, \textit{PSS}, \textit{ReLU}, \textit{sigmoid}, and \textit{tanh}. \textit{PNN} has one middle layer and the number of hidden neurons (h.n.) is 50, while l.r.= 0.05. Fig.~\ref{fig:functioneval} shows the convergence after 500 iterations using four activation functions (\textit{SS}, \textit{PSS}, \textit{sigmoid}, \textit{ReLU} and \textit{tanh}) respectively. We noticed that \textit{PSS} and \textit{SS} have a stable rate of ranking convergence compared to \textit{sigmoid}, \textit{tanh}, and \textit{ReLU}. This stability is due to the stairstep width, which leads each point to reach the correct ranking during \textit{FF} and \textit{BP} in fewer epochs.

\begin{figure}[ht]
\pgfplotsset{footnotesize}
\subfloat[Ranking all labels] {
\begin{tikzpicture}[scale=0.9]
	\begin{axis}[scaled ticks=true,xlabel=\#iterations,
	y label style={at={(axis description cs:-0.15,.5)},anchor=south},
	ylabel=\textit{spearman} $\rho$, grid=major,legend style={at={(1.2,0.6)}}]
\addplot [color=RoyalPurple,mark=*] coordinates { (0,0.7964) (40,0.8305) (80,0.8350) (120,0.8619) (160,0.8733) 
 (200,0.8775) (240,0.8869)  (280,0.8941) (320,0.9161) 
(360,0.8911)  (400,0.9097) (440,0.9047)  (480,0.8994) };
   \addlegendentry{SS} 
    
    \addplot [color=Maroon,mark=+] coordinates { (0,0.44583)  (40,0.85416)  (80,0.8666)  (120,0.866)  (160,0.8791) 
 (200,0.8833) (240,0.8833) (280,0.8833)  (320,0.8833)  
(360,0.8833)  (400,0.8833) (440,0.8833)  (480,0.8833) };
     \addlegendentry{PSS}
    
    \addplot [color=red,mark=square*] coordinates { (0,0.7917)  (40,0.8292)  (80,0.8125) 
    (120,0.8083)(160,0.8208)
    (200,0.8250)(240,0.8292)(280,0.8292)(320,0.8292)(360,0.8208)(400,0.8250)(420,0.8250)(440,0.8208)(480,0.8208)};

      \addlegendentry{Relu}
    \addplot [color=black,mark=*] coordinates { (0,0.1208)  (40,0.1792)  (80,0.3167) 
    (120,0.4083)(160,0.5167)
    (200,0.6125)(240,0.6833)(280,0.7583)(320,0.7750)(360,0.7958)(400,0.8)(440,0.8083)(480,0.8292)};
    \addlegendentry{Sigmoid}
        \addplot [color=ForestGreen,mark=triangle*] coordinates { (0,0.7792)  (40,0.7542)  (80,0.7417) 
    (120,0.7417)(160,0.7250)
    (200,0.7250)(240,0.7292)(280,0.7292)(320,0.7292)(360,0.7208)(400,0.7208)(440,0.7208)(480,0.7208)};
    \addlegendentry{Tanh}
	\end{axis}
\end{tikzpicture}
}
\subfloat[60\% missing labels] {
\begin{tikzpicture}[scale=0.9]
	\begin{axis}[xlabel=\#iterations,
        grid=major,
        legend style={at={(0.75,0.04)},anchor=west}]
  \addplot [color=RoyalPurple,mark=*] coordinates { (0,0.09025105849101828)  (40,0.57848)(80,0.63630338)  (120,0.61323714)  (160,0.6332531) 
 (200,0.64962068)  (240,0.6699358)  (280,0.715470053)  (320, 0.6982852) 
(360,0.7055021)  (400,0.6721687) (440,0.7401709) (480,0.68853629) };

    \addplot [color=Orange,mark=+] coordinates { (0,0.1833333)  (40, 0.4375)  (80,0.4375)  (120, 0.4333) (160,0.4375) 
 (200, 0.4375)  (240,0.4375)  (280,0.4375)  (320, 0.43333)  
(360,0.4416)  (400,0.4583)  (440,0.4583) (480,0.4583) };

    \addplot [color=red,mark=square*]  coordinates { (0,0.09948) (40,0.59583) (80,0.60416) (120,0.629166) (160,0.6208333) 
 (200,0.612)  (240,0.60833)  (280,0.6166)(320, 0.62083) 
(360,0.61971)  (400, 0.61138354) (440,0.6113835) (480,0.60) };
    \addplot [color=black,mark=*]  coordinates { (0,0.03333)  (40,0.295833)  (80, 0.3416666) (120, 0.35)  (160,0.341666) 
 (200,0.3291666)  (240,0.341666) (280,0.35)  (320,0.3625)  
(360, 0.37083) (400,0.383333)  (440, 0.3875) (480,0.385) };

        \addplot [color=ForestGreen,mark=triangle*]  coordinates{ (0,0.41666)  (40, 0.0577350)  (80, 0.0666) (120, 0.0)  (160,0.0) 
(180,0.0)  (220,0.0)  (260,0.05833) (300,0.0)  (340, 0.0) 
(360, 0.0)  (400,0.0)  (440, 0.0) (480,0.0) };

	\end{axis}
\end{tikzpicture}
}
\caption{\textit{PNN} activation function comparison using complete labels and 60\% missing labels in (a) and (b), respectively.}
\label{fig:functioneval}
\end{figure}

\subsubsection{\textit{PSS} and \textit{SS} Evaluation}
As shown in Fig~\ref{fig:functioneval}, \textit{PSS} reaches convergence and remains stable for a long number of iterations compared to \textit{SS}. However, \textit{SS} has better $\rho$ than \textit{PSS}. This good performance of \textit{SS} is due to the reason:
\begin{itemize}
\item The symmetry of \textit{SS} function on the $x$ axis. The \textit{SS} shape handles both positive and negative normalized data. It reduces the number of iterations to reach the correct ranking values.
\end{itemize}
To have the same performance for \textit{SS} and \textit{PSS}, the input data should be scaled from 0 to step width X \#steps and from -$b$ to $b$ for \textit{PSS} and \textit{SS} respectively. 

\subsubsection{Missing Labels Evaluation}
Activation functions are evaluated by removing a random number of labels per instance. \textit{PNN} marked the missing label as -1; \textit{PNN} neglects error calculation during \textit{BP}, $\delta=0$. Thus, the missing label weights remain constants per learning iteration. The missing label approach is applied to the data set by 20\% and 60\% of the training data. The ranking performance decreases when the number of missing labels increases. However, \textit{SS} and \text{PSS} have more stable convergence than other functions. This evaluation is performed on the iris data set, as shown in Fig.~\ref{fig:functioneval}.

\subsubsection{Statistical Test}
\begin{figure}[]
\begin{center}
\includegraphics[scale=0.7]{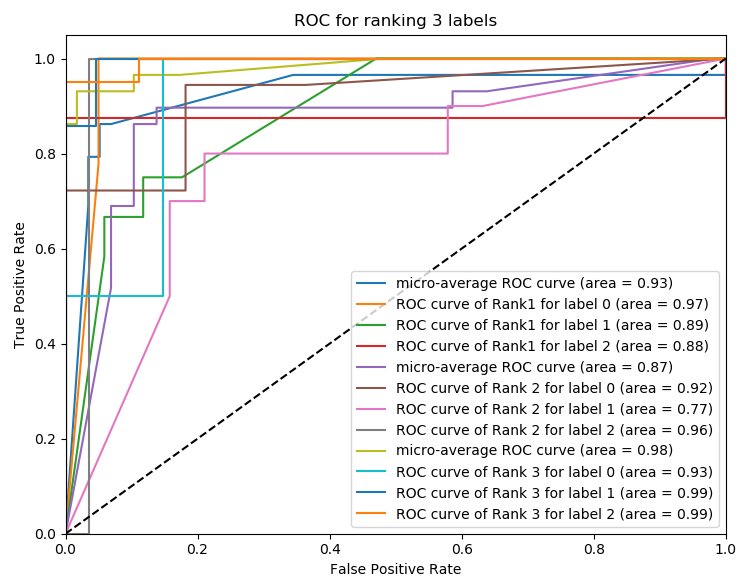}
\caption{ROC of three labels ranking on the wine data set using \textit{PNN} h.n=100 and 50 epochs.}
\label{roc}
\end{center}
\end{figure}

\begin{figure}[ht]

\begin{center}
\includegraphics[scale=0.435]{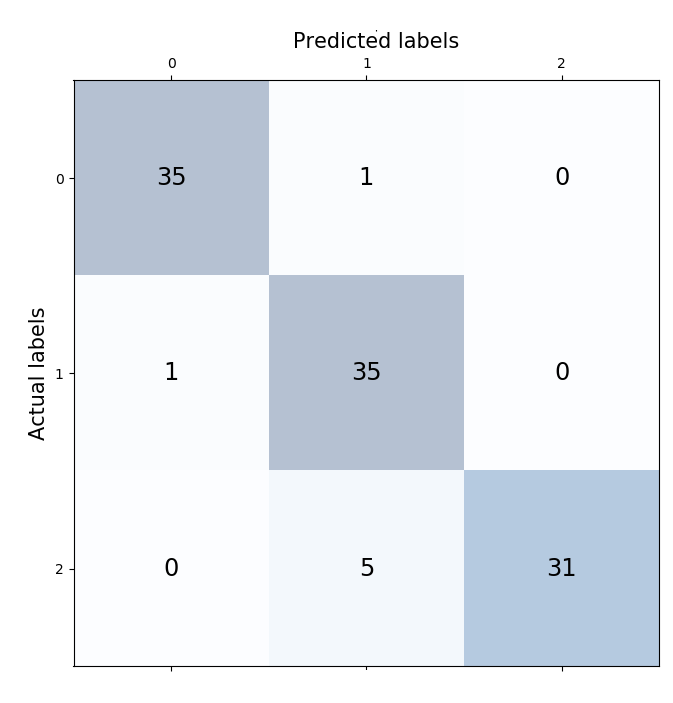}

\footnotesize
\begin{tabular}{|c|c|c|c|}
\hline
Precision& 0.972& 0.85&1.0\\\hline
Recall&0.972&0.972&0.861\\\hline
F1 Score&0.972&0.909&0.925\\\hline
\end{tabular}
\\
\begin{center}
(a)
\end{center}
\end{center}

\begin{center}
\includegraphics[scale=0.65]{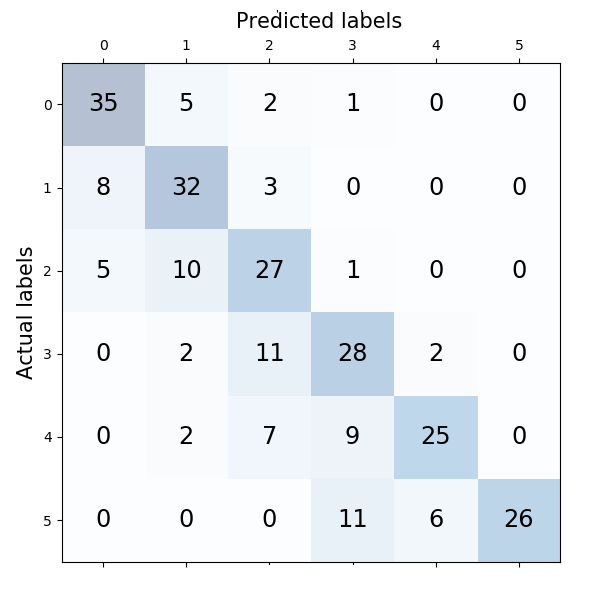}
\footnotesize
\begin{tabular}{|c|c|c|c|c|c|c|}
\hline
Precision& 0.729& 0.627& 0.54&0.56& 0.757&1.0\\\hline
Recall&0.813& 0.744&0.627& 0.651& 0.581& 0.604\\\hline
F1 Score&0.769& 0.680& 0.580& 0.602& 0.657& 0.753\\\hline
\end{tabular}
\end{center}

\begin{center}
(b)
\end{center}

\caption{The confusion matrix of testing the wine, glass data sets where $\tau$ = 0.947, 0.84, Accuracy = 0.935 and 0.8 in (a) and (b) respectively.}
\label{conf}

\end{figure}

The \textit{PNN} results were evaluated using receiver operating characteristic (\textit{ROC}) curves. The true positive and negative for each rank are evaluated per label of wine dataset as shown in Fig.~\ref{roc}. The confusion matrix on wine and glass DS are shown in Fig.~\ref{conf} where $\tau$ = 0.947, 0.84, Accuracy = 0.935 and 0.8 in (a) and (b) respectively.

\subsubsection{Dropout Regularization}
Dropout is applied as a regularization approach to enhance the \textit{PNN} ranking stability by reducing over-fitting. We drop out the weights that have a probability of less than 0.5. these dropped weights are removed from FF, BP, and UW steps. The comparison between dropout and non-dropout of \textit{PNN} are shown in Fig.~\ref{a:droptypeA}. The gap between the training model and ten-fold cross-validation curves has been reduced using dropout regularization using hyperparameters (l.r.=0.05, h.n.=100) on the iris data set. The dropout technique is used with all the data ranking results in the next section.

\begin{figure}[ht]
\pgfplotsset{footnotesize}
\subfloat[Without dropout] {
\begin{tikzpicture}[scale=1]
	\begin{axis}[xlabel=\#iterations,
	y label style={at={(axis description cs:-0.15,.5)},anchor=south},
	ylabel=\textit{spearman} $\rho$,
        grid=major,
        legend style={at={(0.15,0.2)},anchor=west}]
	\addplot coordinates { (0,0.81100423396) (10,0.8110042) (20,0.7693375) (30,0.769337) 
    (40,0.8110)(50,0.769337)(60,0.76933756)(70,0.7693375)
    (80,0.811004)(90,0.7693375)};

	\addplot coordinates { (0,0.6299038) (10,0.8663675134) (20,0.922767090) (30,0.92805021) 
    (40,0.9344497)(50,0.93861645)(60,0.9386164)(70,0.9458333)
    (80,0.946949)(90,0.9386164)};

	\end{axis}
\end{tikzpicture}
}
\subfloat[With dropout] {
\begin{tikzpicture}[scale=1]
	\begin{axis}[xlabel=\#iterations,
        grid=major,
        legend style={at={(0.24,0.22)},anchor=west}]
	\addplot coordinates { (0,0.735843918243516) (10,0.7053418012614796) (20,0.7775105849101829) (30,0.735843918243516) 
    (40,0.6941772515768494)(50,0.69417725)(60,0.622008)(70,0.6150635094)
    (80,0.6941772515)(90,0.6941772515)};
    \addlegendentry{\textit{Pred.}}
	\addplot coordinates { (0,0.2510362971081844) (10,0.6909882023920171) (20,0.8284882023920173) (30,0.8249198421397212) 
    (40,0.7915865088063879)(50,0.8396527520766472)(60,0.8310202655361286)(70,0.8335523286802401)
    (80,0.8188194187433141)(90,0.7879379906938134)};
\addlegendentry{\textit{10-fold. avg.}}

	\end{axis}
\end{tikzpicture}
}
\caption{Training and validation performance without and with dropout regulation approach in (a) and (b) respectively.}
\label{a:droptypeA}
\end{figure}

The following section is the evaluation of ranking experiments using label benchmark data sets.
\section{Experiments}
This section describes the classification and label ranking benchmark data sets, the results using \textit{PN} and \textit{PNN}, and a comparison with existing classification and ranking methods.
\subsection{Data sets}
\subsubsection{Image Classification Data sets}
\textit{PN} is evaluated using \textit{CFAR-100}~\cite{cfar} and Fashion-MNIST~\cite{mnistfashion} data sets. 

\subsubsection{Label Ranking Data sets}
\textit{PNN} is experimented with using three different types of benchmark data sets to evaluate the multi-label ranking performance. The first type of data set focuses on exception preference mining~\cite{newdataset}, and the `algae' data set is the first type that highlights the indifference preferences problem, where labels have repeated preference value~\cite{dataset1}. German elections 2005, 2009, and modified sushi are considered new and restricted preference data sets. The second type is real-world data related to biological science~\cite{Hullermeier}. The third type of data set is semi-synthetic (\textit{SS}) taken from the \textit{KEBI} Data Repository at the Philipps University of Marburg~\cite{Cheng}. All data sets do not have ranking ground truth, and all labels have a continuous permutation space of relations between labels. Table \Romannum{2} summarizes the main characteristics of the data sets.

\begin {table}[]
\caption {Benchmark data sets for label ranking; preference mining~\cite{dataset1}, real-world data sets~\cite{Grbovic} and semi-synthetic (\textit{s-s})~\cite{Cheng}.}~\label{tab:title}
\begin{center}
\begin{tabular}{|c|c|c|c|c|c|}
\hline
\bf Type&\bf DS&\bf Cat.&\bf \#Inst.&\bf \#Attr.&\bf \#lbl.\\ \hline

\multirow{5}{*}{\rotatebox[origin=c]{90}{\parbox[c]{0.7cm}{\centering Mining}}} &algae&chemical stat.&317&11&7\\
&german.2005&user pref.&413 & 31 &5\\
&german.2009&user pref.&413 & 31 & 5\\
&sushi&user pref.&5000&13&7\\
&top7movies&user pref.&602&7&7\\ \hline
\multirow{4}{*}{\rotatebox[origin=c]{90}{\parbox[c]{1.7cm}{\centering Real}}} & cold&biology&2,465&24&4 \\
&diau&biology&2,465&24&7\\
&dtt&biology&2,465&24&4\\
&heat&biology&2,465&24&6\\
&spo&biology&2,465&24&11\\ \hline
\multirow{16}{*}{\rotatebox[origin=c]{90}{\parbox[c]{3.5cm}{\centering Semi-Synthesized}}} &authorship&A&841&70&4\\
&bodyfat&B&252&7&7\\
&calhousing&B&20,640&4&4\\
&cpu-small&B&8192&6&5\\
&elevators&B&16,599&9&9\\
&fried&B&40,769&9&5\\
&glass&A&214&9&6\\
&housing&B&506&6&6\\
&iris&A&150&4&3\\
&pendigits&A&10,992&16&10\\
&segment&A&2310&18&7\\
&stock&B&950&5&5\\
&vehicle&A&846&18&4\\
&vowel&A&528&10&11\\
&wine&A&178&13&3\\
&wisconsin&B&194&16&16\\ \hline
\end{tabular}
\end{center}
\end{table}

\subsection{Results}
\subsubsection{Image Classification Results}
\textit{PN} has 3 kernel sizes of 5,10 and 20 and is tested on the \textit{CFAR-100}~\cite{cfar} data set and 1 kernel with a size 5 for Fashion-MNIST data set~\cite{mnistfashion}. Table \Romannum{3} shows the results compared to other convolutions networks. 
\begin{table}[ht]
\caption {Comparison of classification on CIFAR-100~\cite{cfar} and Fashion-Mnist data set~\cite{mnistfashion} using different convolution models~\label{tab:title1}}
\begin{center}
\begin{tabular}{|c|c|c|c|}
\hline
\bf DS&\bf Model&\bf Baseline&\bf MixUp\\ \hline

\multirow{7}{*}{\rotatebox[origin=c]{90}{{\centering CIFAR-100}}} &ResNet~\cite{resnet}&72.22&78.9\\
&WRN~\cite{wrn}&78.26&82.5 \\
&Dense~\cite{dens}&81.73&83.23 \\ 
&EfficientNetV2-M~\cite{Tan2021EfficientNetV2SM}&92.2&-\\
&EffNet-L2 (SAM)~\cite{mlcoder}&96.08&-\\
&CvT~\cite{Wu2021CvTIC}&94.39&-\\
&PrefNet&80.6&- \\ \hline\hline
\multirow{8}{*}{\rotatebox[origin=c]{90}{{\centering Fashion-MNIST}}} &MLP&0.871&-\\
&RandomForest&0.873&-\\
&LogisticRegression&0.842&-\\ 
&SVC&0.897&-\\
&SGDClassifier&0.81&-\\
&LSTM~\cite{Zhang2018LSTMAI}&0.8757&-\\
&DART~\cite{Tanveer2021FineTuningDF}&0.965&-\\
&PrefNet&0.91&- \\ \hline
\end{tabular}
\end{center}
\end{table}

\subsubsection{Label Ranking Results}
 \textit{PNN} is evaluated by restricted and non-restricted label ranking data sets. The results are derived using \textit{spearman} $\rho$ and converted to \textit{Kendall} $\tau$ coefficient for comparison with other approaches. For data validation, we used 10-fold cross-validation. To avoid the over-fitting problem, We used hyperparameters, i.e. l.r.= (0.0008,0.0005,0.005, 0.05, 0.1) hidden neuron = no.inputs+(5, 10, 50, 100, 200,300,400,450) neurons and scaling boundaries from 1 to 250) are chosen within each cross-validation fold by using the best l.r. on each fold and calculating the average $\tau$ of ten folds. Grid searching is used to obtain the best hyperparameter. For type \textit{B}, we use three output groups and l.r.=0.001 and $w_{b}=0.01$.
 
\subsubsection{Benchmark Results}
Table \Romannum{4} summarizes \textit{PNN} ranking performance of 16 strict label ranking data sets by l.r. and m.n. The results are compared with the four methods for label ranking; supervised clustering~\cite{Grbovic}, supervised decision tree~\cite{Cheng}, \textit{MLP} label ranking~\cite{Ribeiro}, and label ranking tree forest (\textit{LRT})~\cite{forest}. Each method's results are generated by ten-fold cross-validation. The comparison selects only the best approach for each method.

 During the experiment, it was found that ranking performance increases by increasing the number of central neurons up to a maximum of 20 times the number of features. As shown in Table \Romannum{6}, The real datasets are ranked using \textit{PNN} with dropout regulation due to complexity and over-fitting. The dropout requires increasing the number of epochs to reach high accuracy. All the results are held using a single hidden layer with various hidden neurons (100 to 450) and \textit{SS} activation function. The Kendall $\tau$ error converges and reaches close to 1 after 2000 iterations, as shown in Fig.~\ref{comparisonchart}.

\begin{figure*}
\begin{center}
\begin{tikzpicture}[scale=0.8]
\begin{axis}[
title={Comparison of \textit{PNN} with other methods.},
x tick label style={rotate=45, anchor=east},
y tick label style={rotate=45, anchor=east},
symbolic x coords={authorship,bodyfat,callhousing,cpu-small,elevators,fried,glass,housing,iris,pendigits,segment,stock,vehicle,vowel,wine,wisconsin},
enlargelimits=true,
xtick=data,
height=8cm,
width=23cm,
grid=major,
mark size=3.5pt,
scatter/classes={
Diamond={mark=diamond*},
HDiamond={mark=halfdiamond*},
Square={mark=square*},
Triangle={mark=triangle*}
},
ylabel={Kendall $\tau$},
legend style={at={(0.265,0.95)},anchor=east}
]

\addplot [color=RoyalPurple,mark=*] coordinates {
(authorship,0.854)(bodyfat,0.09)(callhousing,0.28)(cpu-small, 0.274) (elevators,0.332)(fried,0.176)
(glass,0.766)(housing,0.246)(iris,0.814)(pendigits,0.422)(segment,0.572)(stock,0.566)
(vehicle,0.738)(vowel,0.49)
(wine,0.898)(wisconsin,0.898)
};

\addplot [mark=triangle*,color=applegreen] coordinates {
(authorship,0.936)(bodyfat,0.281)(callhousing,0.351)(cpu-small,0.5)(elevators,0.768)(fried,0.99)
(glass,0.883)(housing,0.797)(iris,0.966)(pendigits,0.944)(segment,0.959)(stock,0.927)(vehicle,0.862)(vowel,0.9)(wine,0.949)(wisconsin,0.629)
};

\addplot [color=red,mark=square*] coordinates {
(authorship,0.889) (bodyfat,0.075)(callhousing,0.13) (cpu-small,0.357) (elevators,0.687)(fried,0.66)
(glass,0.818)(housing,0.574)(iris,0.911)(pendigits,0.752)(segment,0.842)(stock,0.745)(vehicle,0.801)(vowel,0.545)(wine,0.931)(wisconsin,0.235)
};

\addplot [color=ForestGreen,mark=triangle*] coordinates {
(authorship,0.882)(bodyfat,0.117)(callhousing,0.324) (cpu-small,0.447) (elevators,0.760)(fried,0.890)
(glass,0.883)(housing,0.797)(iris,0.947)(pendigits,0.935)(segment,0.949)(stock,0.895)(vehicle,0.827)(vowel,0.794)(wine,0.882)(wisconsin,0.343)
};

\addplot [mark=diamond*,color = brown] coordinates {
(authorship,0.918) (bodyfat,0.55) (callhousing,0.34) (cpu-small,0.46) (elevators,0.73)(fried,0.91)
(glass,0.8175)(housing,0.712)(iris,0.917)(pendigits,0.86)(segment,0.916)(stock,0.834)(vehicle,0.754)(vowel,0.85)(wine,0.90)(wisconsin,0.612)
};

\legend{$S.Clustering$,$Decision Tree$,$MLP Ranker$,$LRT$,$PNN$}
\end{axis}
\end{tikzpicture}
\caption{Ranking performance comparison of \textit{PNN} with other approaches.}
\label{comparisonchart}
\end{center}
\end{figure*}
Table \Romannum{4} compares \textit{PNN} with similar approaches used for label ranking. These approaches are; Decision trees~\cite{Grbovic}, \textit{MLP-LR}~\cite{Ribeiro} and label ranking trees forest \textit{LRT}~\cite{forest}. In this comparison, we choose the method that has the best results for each approach.
\begin {table}[ht]
\caption {\textit{PNN} performance comparison with various approaches: supervised clustering~\cite{Grbovic}, supervised decision tree~\cite{Cheng}, \textit{MLP} label ranking~\cite{Ribeiro} and label ranking tree forest (\textit{LRT})~\cite{forest}} \label{tab:title2}
\begin{center}
\centering
\addtolength{\tabcolsep}{-2pt}
\begin{tabular}{|c|c|c|c|c|c|c|c|c|c|c|}
\hline
\multicolumn{6}{|c|}{Label Ranking Methods}\\
\hline
\bf{DS}&\bf{S.Clust.}&\bf\textit{DT}&\bf \textit{MLP-LR}&\bf\textit{LRT}&\bf\textit{PNN}\\
\hline
authorship&0.854&0.936(IBLR)&0.889(LA)&0.882&0.918\\\hline
bodyfat&0.09&0.281(CC)&0.075(CA)&0.117&0.5591\\\hline
calhousing&0.28&0.351(IBLR)&0.130(SSGA)&0.324&0.34\\\hline
cpu-small&0.274&0.50(IBLR)&0.357(CA)&0.447&0.46\\\hline
elevators&0.332&0.768(CC)&0.687(LA)&0.760&0.73\\\hline
fried&0.176&0.99(CC)&0.660(CA)&0.890&0.91\\\hline
glass&0.766&0.883(LRT)&0.818(LA)&0.883&0.8175\\\hline
housing&0.246&0.797(LRT)&0.574(CA)&0.797&0.712\\\hline
iris&0.814&0.966(IBLR)&0.911(LA)&0.947&0.917\\\hline
pendigits&0.422&0.944(IBLR)&0.752(CA)&0.935&0.86\\\hline
segment&0.572&0.959(IBLR)&0.842(CA)&0.949&0.916\\\hline
stock&0.566&0.927(IBLR)&0.745(CA)&0.895&0.834\\\hline
vehicle&0.738&0.862(IBLR)&0.801(LA)&0.827&0.754\\\hline
vowel&0.49&0.90(IBLR)&0.545(CA)&0.794&0.85\\\hline
wine&0.898&0.949(IBLR)&0.931(LA)&0.882&0.90\\\hline
wisconsin&0.09&0.629(CC)&0.235(CA)&0.343&0.612\\\hline
\hline
Average&0.475&0.79&0.621&0.730&0.755\\\hline
\end{tabular}
\end{center}
\end{table}
\subsubsection{Preference Mining Results}
The ranking performance of the new preference mining data set is represented in table \Romannum{2}. Two hundred fifty hidden neurons are used To enhance the ranking performance of the algae data set's repeated label values. However, restricted labels ranking data sets of the same type, i.e., (German elections and sushi), did not require a high number of hidden neurons and incurred less computational cost.


Experiments on the real-world biological data set were conducted using supervised clustering (\textit{SC})~\cite{Grbovic}, Table \Romannum{5} presents the comparison between \textit{PNN} and supervised clustering on biological real world data in terms of {$Loss_{LR}$} as given in Eq.~\ref{tab:loss1}.

\begin{equation}\label{tab:loss1}
\tau= 1-2\cdot Loss_{LR}
\end{equation}
where $\tau$ is Kendall $\tau$ ranking error and $Loss_{LR}$ is the ranking loss function.

\textit{SS} function with 16 steps is used to rank Wisconsin data set with 16 labels. By increasing the number of steps in the interval and scaling up the features between -100 and 100, The step width is small. To enhance ranking performance, the data set has many labels. The number of hidden neurons is increased to exceed $\tau=0.5$.

\begin {table}[ht]
\caption {Comparison between \textit{PNN} and supervised clustered on biological real world data in terms of {$Loss_{LR}$}} \label{tab:title4}
\begin{center}
\begin{tabular}{ |c|c|c|c|c|c| }
\hline
\multicolumn{3}{|c|}{Biological real world data} \\
\hline
\bf{DS}&\bf{S.Clustering}&\bf{\textit{PNN}} \\ \hline
cold & 0.198 & 0.11\\ \hline
diau &0.304 & 0.255\\ \hline
dtt & 0.124& 0.01\\ \hline
heat & 0.072 &0.013 \\ \hline
spo & 0.118 & 0.014\\ \hline
\hline
Average & 0.1632 & 0.0804\\ \hline
\end{tabular}
\end{center}
\end{table}

\begin {table*}
\begin{center}
\caption {\textit{PNN} label ranking performance in terms of $\tau$ coefficient, learning step and the number of middle layer neurons (\#m.n). The training per fold and testing time is given in the last two columns. ‘s’, ‘m’, ‘h’ denote seconds, minutes and hours, respectively.}
\label{tab:title5}
\begin{tabular}{|c|c|c|c|c|c|c|c|c|c|c|}\hline
\bf Type& \bf DS& \bf Avg. $\tau$ &\bf \#m.n.&\bf l.r. &\bf \#Iterations.&\bf Dropout &\bf Scaling. &\bf Training t.& \bf Testing t.\\ \cline{1-10}

\multirow{7}{*}{\rotatebox[origin=c]{90}{Real}}
& cold &0.4 &10 &0.0008&2000&yes&-4:4 & 2.8h&1.2s\\\cline{2-10}

    & diau &0.466  &400 &0.0005&2500&yes&-2:2 &2.9h &4s \\\cline{2-10}
    & dtt &0.60&400 &0.0001&5000&yes&-4:4 &5.7h &1.88s\\\cline{2-10}

       & heat &0.876&450&0.0005&5000&yes&-2:2 &6.2h &1.18s\\\cline{2-10}

       &spo &0.8  &300 &0.0005&5000&yes&-4:4 &7.4h &0.98s\\\cline{2-10}
       &German2005&0.8  &300 &0.0005&1000&no&-4:4 &35.15m &0.0879s\\\cline{2-10}
            &German2009&0.67  &300&0.0005&500&no&-4:4 &7.087m &0.105s\\\cline{2-10}
\hline
\multirow{15}{*}{\rotatebox[origin=c]{90}{\parbox[c]{3.5cm}{\centering Semi-Synthesized}}}

       &authorship&0.931 &200 &0.0008&200&no&-4:4 &3.82m &0.34s\\\cline{2-10}

       &bodyfat &0.559 &100 &0.0005&2500&yes&-2:2 &16.92m &0.44s\\\cline{2-10}

       &calhousing &0.34&200&0.0007&1000&no&-2:2 &5.03h &4.127s\\\cline{2-10}

       &cpu-small&0.46 &200&0.005&1000&no&-2:2 &2.089h &1.717\\\cline{2-10}

       &elevators &0.73 &20&0.003&100&no&-2:2 &27.03m &3.7s\\\cline{2-10}

       &fried &0.89 &100&0.005&100&no&-2:2 &1.02h &8.45s\\\cline{2-10}
       &glass&0.948 &100 &0.005&100&no&-3:3 &14.8s &0.04s\\\cline{2-10}
       &housing &0.7615&25&0.005&100&no&-3:3 &37.21s &0.1s\\\cline{2-10}
           &iris&0.956&100&0.005&100&no&-3:3 &29.39s &0.066s\\\cline{2-10}
       &pendigits &0.86&100&0.005&100&no&-3:3 &34.6m &5.69s\\\cline{2-10}
       &segment & 0.956 &20&0.007&100&no&-3:3 & 440.8s&0.94s\\\cline{2-10}
            &stock &0.868 &100&0.005&100&no&-3:3 &142.48s &0.87s\\\cline{2-10}
            &vehicle&0.869 &100&0.005&100&no&-3:3 &91s &0.2s\\\cline{2-10}
       &vowel& 0.85  &100&0.005&100 &no&-3:3 &88.37s &0.312s\\\cline{2-10}
       &wine & 0.90 &100&0.005&100&no&-3:3 &19.19s &0.063s\\\cline{2-10}
       &wisconsin & 0.61 &300&0.0005&2500&yes&-4:4 &13.56m &0.1332s\\\cline{1-10}

\end{tabular}
\end{center}
\end{table*}
\subsection{Computational Platform}
\textit{PNN} and \textit{PN} is implemented from scratch without the Tensorflow API and developed using Numba API to speed the execution on the GPU and use Cuda 10.1 and Tensorflow-GPU 2.3 for GPU execution and executed at the University of Technology Sydney High-Performance Computing cluster based on Linux RedHat 7.7, which has an NVIDIA Quadro GV100 and memory of 32 G.B. For a non-GPU version of \textit{PNN} is located at GitHub Repository~\cite{pnngithub}.

\subsection{Discussion and Future Work}
It can be noticed from table \Romannum{2} that \textit{PN} is performing better than ResNet~\cite{resnet} and WRN~\cite{wrn}.
Different types of architectures of \textit{PN} could be used to enhance the results and reach state-of-the-art in terms of image classification~\cite{10.1007/978-3-030-34879-3_2,8615011,8258754}.
It can be noticed from table \Romannum{3} that \textit{PNN} outperforms on \textit{SS} data sets with $\tau_{Avg}=0.8$, whereas other methods such as, supervised clustering, decision tree, \textit{MLP-ranker} and \textit{LRT}, have results $\tau_{Avg}= 0.79, 0.73, 0.62, 0.475,$ respectively.
Also, the performance of \textit{PNN} is almost 50\% better than supervised clustering in terms of ranking loss function $Loss_{LR}$ on real-world biological data set, as shown in table \Romannum{5}.
The superiority of \textit{PNN} is used for classification and ranking problems. The ranking is used in input data as a feature selection criteria is a novel approach for deep learning.

Encoding the labels' preference relation to numeric values and ranking the output labels simultaneously in one model is an advanced step over pairwise label ranking based on classification. \textit{PNN} could be used to solve new preference mining problems. One of these problems is incomparability between labels, where Label ranking has incomparable relation $\bot$, i.e., ranking space ($\lambda_{a}\succ\lambda_{b}\bot\lambda_{c}$) is encoded to (1, 2, -1) and $(\lambda_{a}\succ\lambda_{b})\bot(\lambda_{c}\succ\lambda_{d})$ is encoded to (1, 2, -1, -2). \textit{PNN} could be used to solve new problem of non-strict partial orders ranking, i.e., ranking space ($\lambda_{a}\succ\lambda_{b}\succeq\lambda_{c}$) is encoded to (1, 2, 3) or (1, 2, 2). Future research may enhance \textit{PN} by adding kernel size and \textit{SS} parameters as part of the deep learning to choose the best kernel size and \textit{SS} step width, which could enhance the image attention. Modifying \textit{PNN} architecture by adding bias and solving noisy label ranking problems.

\section{Conclusion}
This paper proposed a novel method to rank a complete multi-label space in output labels and features extraction in both simple and deep learning.\textit{PN} is a new research direction for image recognition based on new kernel and pixel calculations. \textit{PNN} and \textit{PN} are native ranker networks for image classification and label ranking problems that uses \textit{SS} or \textit{PSS} to rank the multi-label per instance. This neural network's novelty is a new kernel mechanism, activation, and objective functions. This approach takes less computational time with a single middle layer. It is indexing multi-labels as output neurons with preference values. The neuron output structure can be mapped to integer ranking value; thus, \textit{PNN} accelerates the ranking learning by assigning the rank value to more than one output layer to reinforce updating the random weights. \textit{PNN} is implemented using python programming language 3.6~\cite{pnngithub}, and activation functions are modelled using wolframe Mathematica software~\cite{Wolfram}. A video demo that shows the ranking learning process using toy data is available to download~\cite{videofile}.

\section*{Acknowledgement}
This work was supported in part by the Australian Research Council (ARC) under discovery grants DP180100656 and DP210101093. The research was also sponsored in part by the US Office of Naval Research Global under Cooperative Agreement Number ONRG - NICOP - N62909-19-1-2058 and AFOSR – DST Australian Autonomy Initiative agreement ID10134. We also thank the NSW Defence Innovation Network and NSW State Government of Australia for financial support in part of this research through grant PP21-22.03.02.
\bibliographystyle{IEEEtran}
\bibliography{main}

\vskip 0pt plus -1fil
\begin{IEEEbiography}
[{\includegraphics[width=1in,height=1.25in,clip,keepaspectratio]{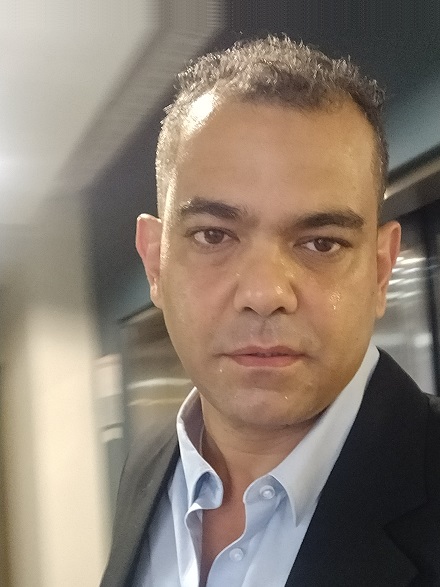}}]{Ayman Elgharabawy}
Dr. Ayman Elgharabawy received a B.S. and master's degree from the Biomedical Engineering Dep., Faculty of Engineering, Cairo University, Egypt, in 2000 and 2015. He received a Computer engineering diploma from Computer Engineering Dep., Faculty of Engineering Ain Shams University, Cairo, Egypt, he worked as a Programmer, software designer and technical architect for 14 years in many IT Companies (HP, CGI-Logica and Vodafone), his master thesis was in Brain-Computer Interface. He finished Ph.D. in 2022 in advanced machine learning at, school of computer science, University of Technology Sydney (UTS). He is a research fellow at the biological data science institute (BDSI) at the college of science, Australian National University (ANU) and visiting scientist at CSIRO A\&F. His research area includes Neural network design (Neuromorphic Computing), preference learning, bioinformatics. graph neural network and image processing.
\end{IEEEbiography}
\vskip 0pt plus -1fil
\begin{IEEEbiography}
[{\includegraphics[width=1in,height=1.25in,clip,keepaspectratio]{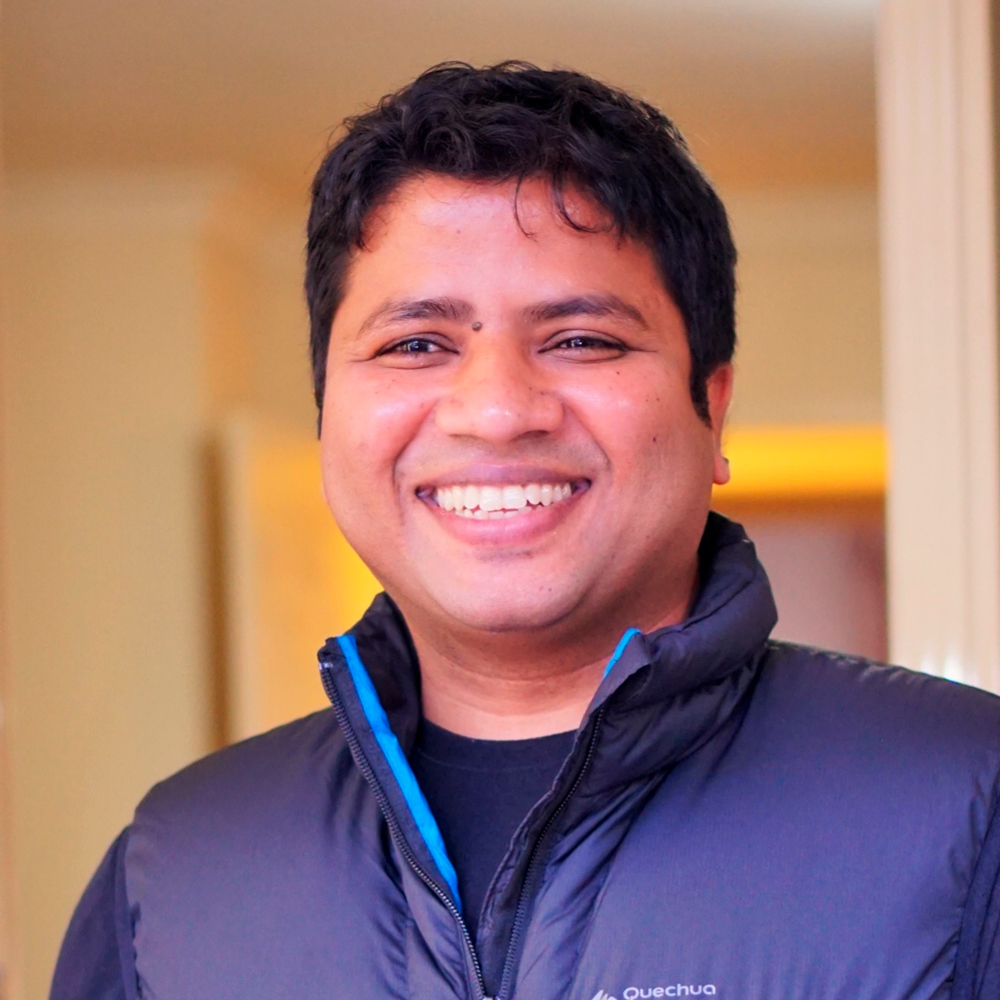}}]{Mukesh Prasad}
Dr. Mukesh Prasad (SMIEEE, ACM) is a Senior Lecturer in the School of Computer Science (SoCS), Faculty of Engineering and Information Technology (FEIT), University of Technology Sydney (UTS), Australia.
His research expertise lies in the development of new methods in artificial intelligence, machine learning and data analytic approach within the domain of computer vision, healthcare, biomedical, internet of things and brain-computer interface and marketing research. He has published more than 100 articles including several prestigious IEEE Transactions and other Top Q1 journals and conferences in the areas of Artificial Intelligence and Machine Learning. 

He received an M.S. degree from the School of Computer Systems and Sciences, Jawaharlal Nehru University, New Delhi, India, in 2009, and a PhD degree from the Department of Computer Science, National Chiao Tung University, Hsinchu, Taiwan, in 2015.
He worked as a principal engineer at Taiwan Semiconductor Manufacturing Company, Hsinchu, Taiwan during 2016-2017.
He started his academic career as a Lecturer at the University of Technology Sydney in 2017. He is also an Associate/Area Editor of several top journals in the field of machine learning, computational intelligence, and emergent technologies.
\end{IEEEbiography}
\vskip 0pt plus -1fil
\begin{IEEEbiography}
[{\includegraphics[width=1in,height=1.25in,clip,keepaspectratio]{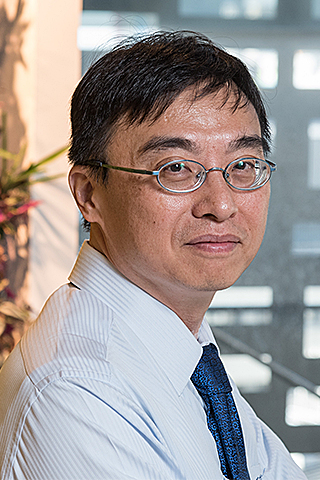}}]{Prof. Chin-Teng-Lin}
Dr. Chin-Teng Lin received a B.S. degree from National Chiao-Tung University (NCTU), Taiwan in 1986, and a Master and PhD degree in electrical engineering from Purdue University, USA in 1989 and 1992, respectively. He is currently the Chair Professor of the Faculty of Engineering and Information Technology, University of Technology Sydney, Chair Professor of Electrical and Computer Engineering, NCTU, International Faculty of the University of California at San Diego (UCSD), and Honorary Professorship of the University of Nottingham. 

Dr. Lin was elevated to an IEEE Fellow for his contributions to biologically inspired information systems in 2005 and was elevated to International Fuzzy Systems Association (IFSA) Fellow in 2012. He has been elected as the Editor-in-chief of IEEE Transactions on Fuzzy Systems since 2011.

Dr. Lin is the Distinguished Lecturer at the IEEE CAS Society from 2003 to 2005, and CIS Society from 2015-2017. He served as the Deputy Editor-in-Chief of IEEE Transactions on Circuits and Systems-II from 2006-2008.

Dr. Lin is the coauthor of Neural Fuzzy Systems (Prentice-Hall), and the author of Neural Fuzzy Control Systems with Structure and Parameter Learning (World Scientific). He has published more than 200 journal papers (Total Citation: 20,155, H-index: 53, i10-index: 373) in neural networks, fuzzy systems, multimedia hardware/software, and cognitive neuro-engineering, including approximately 101 IEEE journal papers.
\end{IEEEbiography}

\end{document}

%% file: TypeA.tikz
\begin{tikzpicture}[scale=0.9,multilayer=2d,rotate=-90]

\Vertex[x=5,y=-1,layer=3,color=lightgray,opacity=1]{F6}
\Vertex[x=4,y=-1,layer=3,color=lightgray,opacity=1]{F5}
\Vertex[x=3,y=-1,layer=3,color=lightgray,opacity=1]{F4}
\Vertex[x=2,y=-1,layer=3,color=lightgray,opacity=1]{F3}
\Vertex[x=1.,y=-1,layer=3,color=lightgray,opacity=1]{F2}
\Vertex[x=0,y=-1,layer=3,color=lightgray,opacity=1]{F1}

\Vertex[x=7,y=1,layer=3,color=lightgray,opacity=1]{h9}
\Vertex[x=6,y=1,layer=3,color=lightgray,opacity=1]{h8}
\Vertex[x=5,y=1,layer=3,color=lightgray,opacity=1]{h7}
\Vertex[x=4,y=1,layer=3,color=lightgray,opacity=1]{h6}
\Vertex[x=3,y=1,layer=3,color=lightgray,opacity=1]{h5}
\Vertex[x=2,y=1,layer=3,color=lightgray,opacity=1]{h4}
\Vertex[x=1,y=1,layer=3,color=lightgray,opacity=1]{h3}
\Vertex[x=0,y=1,layer=3,color=lightgray,opacity=1]{h2}
\Vertex[x=-1,y=1,layer=3,color=lightgray,opacity=1]{h1}

\Vertex[x=0.5,y=3,layer=2,color=lightgray,opacity=1]{o1}
\Vertex[x=1.5,y=3,layer=2,color=lightgray,opacity=1]{o2}
\Vertex[x=2.5,y=3,layer=2,color=lightgray,opacity=1]{o3}
\Vertex[x=3.5,y=3,layer=2,color=lightgray,opacity=1]{o4}
\Vertex[x=4.5,y=3,layer=2,color=lightgray,opacity=1]{o5}

\begin{Layer}[layer=3]

\draw[fill] (5.6,-1) circle [radius=0.1];
\draw[fill] (5.9,-1) circle [radius=0.07];
\draw[fill] (6.2,-1) circle [radius=0.05];

\draw[fill] (7.6,1) circle [radius=0.1];
\draw[fill] (7.9,1) circle [radius=0.07];
\draw[fill] (8.2,1) circle [radius=0.05];

\draw[fill] (5.2,3) circle [radius=0.1];
\draw[fill] (5.5,3) circle [radius=0.07];
\draw[fill] (5.8,3) circle [radius=0.05];

\node [rotate=90] at (5,-1) {\tiny$F_6$};
\node [rotate=90] at (4,-1) {\tiny$F_5$};
\node [rotate=90] at (3,-1) {\tiny$F_4$};
\node [rotate=90] at (2,-1) {\tiny$F_3$};
\node [rotate=90] at (1,-1) {\tiny$F_2$};
\node [rotate=90] at (0,-1) {\tiny$F_1$};

\node [rotate=90] at (-1,1) {\tiny$H_1$};
\node [rotate=90] at (0,1) {\tiny$H_2$};
\node [rotate=90] at (1,1) {\tiny$H_3$};
\node [rotate=90] at (2,1) {\tiny$H_4$};
\node [rotate=90] at (3,1) {\tiny$H_5$};
\node [rotate=90] at (4,1) {\tiny$H_6$};
\node [rotate=90] at (5,1) {\tiny$H_7$};
\node [rotate=90] at (6,1) {\tiny$H_8$};
\node [rotate=90] at (7,1) {\tiny$H_9$};

\node [rotate=90] at (.5,3) {\tiny$O_1$};
\node [rotate=90] at (1.5,3) {\tiny$O_2$};
\node [rotate=90] at (2.5,3) {\tiny$O_3$};
\node [rotate=90] at (3.5,3) {\tiny$O_4$};
\node [rotate=90] at (4.5,3) {\tiny$O_5$};

\draw[dotted] (-1.5,0) -- (4.5,0);
\end{Layer}
\Edge[Math,distance=0.35,fontscale=0.5,Direct,lw=0.2,bend=0](F1)(h1)
\Edge[Math,distance=0.2,fontscale=0.5,Direct,lw=0.2,bend=0](F1)(h2)
\Edge[Math,distance=0.2,fontscale=0.5,Direct,lw=0.2,bend=0](F1)(h3)
\Edge[Math,distance=0.2,fontscale=0.5,Direct,lw=0.2,bend=0](F1)(h4)
\Edge[Math,distance=0.2,fontscale=0.5,Direct,lw=0.2,bend=0](F1)(h5)
\Edge[Math,distance=0.2,fontscale=0.5,Direct,lw=0.2,bend=0](F1)(h6)
\Edge[Math,distance=0.2,fontscale=0.5,Direct,lw=0.2,bend=0](F1)(h7)
\Edge[Math,distance=0.2,fontscale=0.5,Direct,lw=0.2,bend=0](F1)(h8)
\Edge[Math,distance=0.2,fontscale=0.5,Direct,lw=0.2,bend=0](F1)(h9)

\Edge[Math,distance=0.32,fontscale=0.5,Direct,lw=0.2,bend=0](F2)(h1)
\Edge[Math,distance=0.14,fontscale=0.5,Direct,lw=0.2,bend=0](F2)(h2)
\Edge[Math,distance=0.14,fontscale=0.5,Direct,lw=0.2,bend=0](F2)(h3)
\Edge[Math,distance=0.2,fontscale=0.5,Direct,lw=0.2,bend=0](F2)(h4)
\Edge[Math,distance=0.2,fontscale=0.5,Direct,lw=0.2,bend=0](F2)(h5)
\Edge[Math,distance=0.2,fontscale=0.5,Direct,lw=0.2,bend=0](F2)(h6)
\Edge[Math,distance=0.2,fontscale=0.5,Direct,lw=0.2,bend=0](F2)(h7)
\Edge[Math,distance=0.2,fontscale=0.5,Direct,lw=0.2,bend=0](F2)(h8)
\Edge[Math,distance=0.2,fontscale=0.5,Direct,lw=0.2,bend=0](F2)(h9)

\Edge[Math,distance=0.66,fontscale=0.5,Direct,lw=0.2,bend=0](F3)(h1)
\Edge[Math,distance=0.65,fontscale=0.5,Direct,lw=0.2,bend=0](F3)(h2)
\Edge[Math,distance=0.65,fontscale=0.5,Direct,lw=0.2,bend=0](F3)(h3)
\Edge[Math,distance=0.2,fontscale=0.5,Direct,lw=0.2,bend=0](F3)(h4)
\Edge[Math,distance=0.2,fontscale=0.5,Direct,lw=0.2,bend=0](F3)(h5)
\Edge[Math,distance=0.2,fontscale=0.5,Direct,lw=0.2,bend=0](F3)(h6)
\Edge[Math,distance=0.2,fontscale=0.5,Direct,lw=0.2,bend=0](F3)(h7)
\Edge[Math,distance=0.2,fontscale=0.5,Direct,lw=0.2,bend=0](F3)(h8)
\Edge[Math,distance=0.2,fontscale=0.5,Direct,lw=0.2,bend=0](F3)(h9)

\Edge[Math,distance=0.1,fontscale=0.5,Direct,lw=0.2,bend=0](F4)(h1)
\Edge[Math,distance=0.25,fontscale=0.5,Direct,lw=0.2,bend=0](F4)(h2)
\Edge[Math,distance=0.25,fontscale=0.5,Direct,lw=0.2,bend=0](F4)(h3)
\Edge[Math,distance=0.2,fontscale=0.5,Direct,lw=0.2,bend=0](F4)(h4)
\Edge[Math,distance=0.2,fontscale=0.5,Direct,lw=0.2,bend=0](F4)(h5)
\Edge[Math,distance=0.2,fontscale=0.5,Direct,lw=0.2,bend=0](F4)(h6)
\Edge[Math,distance=0.2,fontscale=0.5,Direct,lw=0.2,bend=0](F4)(h7)
\Edge[Math,distance=0.2,fontscale=0.5,Direct,lw=0.2,bend=0](F4)(h8)
\Edge[Math,distance=0.2,fontscale=0.5,Direct,lw=0.2,bend=0](F4)(h9)

\Edge[Math,distance=0.1,fontscale=0.5,Direct,lw=0.2,bend=0](F5)(h1)
\Edge[Math,distance=0.25,fontscale=0.5,Direct,lw=0.2,bend=0](F5)(h2)
\Edge[Math,distance=0.25,fontscale=0.5,Direct,lw=0.2,bend=0](F5)(h3)
\Edge[Math,distance=0.2,fontscale=0.5,Direct,lw=0.2,bend=0](F5)(h4)
\Edge[Math,distance=0.2,fontscale=0.5,Direct,lw=0.2,bend=0](F5)(h5)
\Edge[Math,distance=0.2,fontscale=0.5,Direct,lw=0.2,bend=0](F5)(h6)
\Edge[Math,distance=0.2,fontscale=0.5,Direct,lw=0.2,bend=0](F5)(h7)
\Edge[Math,distance=0.2,fontscale=0.5,Direct,lw=0.2,bend=0](F5)(h8)
\Edge[Math,distance=0.2,fontscale=0.5,Direct,lw=0.2,bend=0](F5)(h9)

\Edge[Math,distance=0.1,fontscale=0.5,Direct,lw=0.2,bend=0](F6)(h1)
\Edge[Math,distance=0.25,fontscale=0.5,Direct,lw=0.2,bend=0](F6)(h2)
\Edge[Math,distance=0.25,fontscale=0.5,Direct,lw=0.2,bend=0](F6)(h3)
\Edge[Math,distance=0.2,fontscale=0.5,Direct,lw=0.2,bend=0](F6)(h4)
\Edge[Math,distance=0.2,fontscale=0.5,Direct,lw=0.2,bend=0](F6)(h5)
\Edge[Math,distance=0.2,fontscale=0.5,Direct,lw=0.2,bend=0](F6)(h6)
\Edge[Math,distance=0.2,fontscale=0.5,Direct,lw=0.2,bend=0](F6)(h7)
\Edge[Math,distance=0.2,fontscale=0.5,Direct,lw=0.2,bend=0](F6)(h8)
\Edge[Math,distance=0.2,fontscale=0.5,Direct,lw=0.2,bend=0](F6)(h9)


\Edge[Math,distance=0.21,fontscale=0.5,lw=0.2,Direct,bend=0](h1)(o1)
\Edge[Math,distance=0.25,fontscale=0.5,lw=0.2,Direct,bend=0](h1)(o2)
\Edge[Math,distance=0.25,fontscale=0.5,lw=0.2,Direct,bend=0](h1)(o3)
\Edge[Math,distance=0.25,fontscale=0.5,lw=0.2,Direct,bend=0](h1)(o4)
\Edge[Math,distance=0.25,fontscale=0.5,lw=0.2,Direct,bend=0](h1)(o5)

\Edge[Math,distance=0.21,fontscale=0.5,lw=0.2,Direct,bend=0](h2)(o1)
\Edge[Math,distance=0.25,fontscale=0.5,lw=0.2,Direct,bend=0](h2)(o2)
\Edge[Math,distance=0.25,fontscale=0.5,lw=0.2,Direct,bend=0](h2)(o3)
\Edge[Math,distance=0.25,fontscale=0.5,lw=0.2,Direct,bend=0](h2)(o4)
\Edge[Math,distance=0.25,fontscale=0.5,lw=0.2,Direct,bend=0](h2)(o5)

\Edge[Math,distance=0.21,fontscale=0.5,lw=0.2,Direct,bend=0](h3)(o1)
\Edge[Math,distance=0.25,fontscale=0.5,lw=0.2,Direct,bend=0](h3)(o2)
\Edge[Math,distance=0.25,fontscale=0.5,lw=0.2,Direct,bend=0](h3)(o3)
\Edge[Math,distance=0.25,fontscale=0.5,lw=0.2,Direct,bend=0](h3)(o4)
\Edge[Math,distance=0.25,fontscale=0.5,lw=0.2,Direct,bend=0](h3)(o5)

\Edge[Math,distance=0.21,fontscale=0.5,lw=0.2,Direct,bend=0](h4)(o1)
\Edge[Math,distance=0.25,fontscale=0.5,lw=0.2,Direct,bend=0](h4)(o2)
\Edge[Math,distance=0.25,fontscale=0.5,lw=0.2,Direct,bend=0](h4)(o3)
\Edge[Math,distance=0.25,fontscale=0.5,lw=0.2,Direct,bend=0](h4)(o4)
\Edge[Math,distance=0.25,fontscale=0.5,lw=0.2,Direct,bend=0](h4)(o5)

\Edge[Math,distance=0.21,fontscale=0.5,lw=0.2,Direct,bend=0](h5)(o1)
\Edge[Math,distance=0.25,fontscale=0.5,lw=0.2,Direct,bend=0](h5)(o2)
\Edge[Math,distance=0.25,fontscale=0.5,lw=0.2,Direct,bend=0](h5)(o3)
\Edge[Math,distance=0.25,fontscale=0.5,lw=0.2,Direct,bend=0](h5)(o4)
\Edge[Math,distance=0.25,fontscale=0.5,lw=0.2,Direct,bend=0](h5)(o5)

\Edge[Math,distance=0.21,fontscale=0.5,lw=0.2,Direct,bend=0](h6)(o1)
\Edge[Math,distance=0.25,fontscale=0.5,lw=0.2,Direct,bend=0](h6)(o2)
\Edge[Math,distance=0.25,fontscale=0.5,lw=0.2,Direct,bend=0](h6)(o3)
\Edge[Math,distance=0.25,fontscale=0.5,lw=0.2,Direct,bend=0](h6)(o4)
\Edge[Math,distance=0.25,fontscale=0.5,lw=0.2,Direct,bend=0](h6)(o5)

\Edge[Math,distance=0.21,fontscale=0.5,lw=0.2,Direct,bend=0](h7)(o1)
\Edge[Math,distance=0.25,fontscale=0.5,lw=0.2,Direct,bend=0](h7)(o2)
\Edge[Math,distance=0.25,fontscale=0.5,lw=0.2,Direct,bend=0](h7)(o3)
\Edge[Math,distance=0.25,fontscale=0.5,lw=0.2,Direct,bend=0](h7)(o4)
\Edge[Math,distance=0.25,fontscale=0.5,lw=0.2,Direct,bend=0](h7)(o5)

\Edge[Math,distance=0.21,fontscale=0.5,lw=0.2,Direct,bend=0](h8)(o1)
\Edge[Math,distance=0.25,fontscale=0.5,lw=0.2,Direct,bend=0](h8)(o2)
\Edge[Math,distance=0.25,fontscale=0.5,lw=0.2,Direct,bend=0](h8)(o3)
\Edge[Math,distance=0.25,fontscale=0.5,lw=0.2,Direct,bend=0](h8)(o4)
\Edge[Math,distance=0.25,fontscale=0.5,lw=0.2,Direct,bend=0](h8)(o5)

\Edge[Math,distance=0.21,fontscale=0.5,lw=0.2,Direct,bend=0](h9)(o1)
\Edge[Math,distance=0.25,fontscale=0.5,lw=0.2,Direct,bend=0](h9)(o2)
\Edge[Math,distance=0.25,fontscale=0.5,lw=0.2,Direct,bend=0](h9)(o3)
\Edge[Math,distance=0.25,fontscale=0.5,lw=0.2,Direct,bend=0](h9)(o4)
\Edge[Math,distance=0.25,fontscale=0.5,lw=0.2,Direct,bend=0](h9)(o5)

\end{tikzpicture}

%% file: grid1.tikz
\newcounter{row}
\newcounter{col}

\newcommand\setrow[6]{
  \setcounter{col}{1}
  \foreach \n in {#1, #2, #3, #4, #5, #6} {
    \edef\x{\value{col} - 0.5}
    \edef\y{6.5 - \value{row}}
    \node[anchor=center] at (\x, \y) {\n};
    \stepcounter{col}
  }
  \stepcounter{row}
}

\begin{tikzpicture}[scale=.5]

  \begin{scope}
    \draw (0, 0) grid (6, 6);
    \foreach \y in {0,1,...,6}
    {\draw (0,\y) -- (6.5,\y);}
    \foreach \x in {0,1,...,6}
    {\draw (\x,0) -- (\x,6.5);}
    
    \setcounter{row}{1}
    \setrow {0}{2}{10 }  {5}{ 95}{1}  
    \setrow {8}{10 }{140 }  {2}{68 }{3}  
    \setrow { 138}{3}{ 255}  {240 }{6}{54 } 
    \setrow {53 }{155 }{1}  {255 }{64 }{195 } 
    \setrow {5}{4}{56}  {167 }{230 }{42 }  
    \setrow { 0}{ 65}{2}  {94 }{ 69}{ 12}

  \end{scope}
  \end{tikzpicture}

%% file: grid2.tikz
\newcommand\setrow[6]{
  \setcounter{col}{1}
  \foreach \n in {#1, #2, #3, #4, #5, #6} {
    \edef\x{\value{col} - 0.5}
    \edef\y{6.5 - \value{row}}
    \node[anchor=center] at (\x, \y) {\n};
    \stepcounter{col}
  }
  \stepcounter{row}
}

\begin{tikzpicture}[scale=.5]

  \begin{scope}
    \draw (0, 0) grid (6, 6);
    \foreach \y in {0,1,...,6}
    {\draw (0,\y) -- (6.5,\y);}
    \foreach \x in {0,1,...,6}
    {\draw (\x,0) -- (\x,6.5);}
    \draw[very thick, scale=1] (0, 0) grid (3, 3);
    \setcounter{row}{1}
    \setrow {1}{3}{9}{6}{20}{2}  
    \setrow {8}{9}{22}{3}{17}{4}  
    \setrow {21}{4}{28}{27}{7}{13} 
    \setrow {12}{23}{2}{28}{15}{25} 
    \setrow {6}{5}{14}{24}{26}{11}  
    \setrow {1}{16}{3}{19}{18}{10}

  \end{scope}
  \end{tikzpicture}

%% file: grid3.tikz
\begin{tikzpicture}[scale=.5]


\newcommand\setrow[3]{
  \setcounter{col}{1}
  \foreach \n in {#1, #2, #3} {
    \edef\x{\value{col} - 0.5}
    \edef\y{3.5 - \value{row}}
    \node[anchor=center] at (\x, \y) {\n};
    \stepcounter{col}
  }
  \stepcounter{row}
}

  \begin{scope}

    \draw[very thick, scale=1] (0, 0) grid (3, 3);
    \setcounter{row}{1}
    \setrow {6}{9}{2}   
    \setrow {5}{4}{7}  
    \setrow {1}{8}{3} 

  \end{scope}
  
\end{tikzpicture}

%% file: flowchart.tikz
\begin{tikzpicture}[node distance = 5mm and 7mm,
     start chain = going right,alg/.style = {draw, align=center},
  circle/.style = {shape=circle, aspect=2.2, draw},font=\linespread{0.8}\selectfont]
    \begin{scope}[every node/.append style={on chain, join=by -Stealth}]
\node (n1) [alg] {\footnotesize {greyscale} \\ \footnotesize {image}};
\node (n2) [alg]  {\footnotesize {flatten} \\\footnotesize{image}};
\node (n3) [alg]  {\footnotesize{pixels}\\\footnotesize{averaging}};
\node (n4) [alg] {\footnotesize{pixels} \\\footnotesize{sorting}};
\node (n5) [alg]  {\footnotesize {image}\\\footnotesize{2D}};
\node (n6) [alg]  {\footnotesize {image}\\\footnotesize{window}\\ \footnotesize{scanning}};
\node (n7) [alg]  {\footnotesize {flatten} \\ \footnotesize{window}};
\node (n8) [alg]  {\footnotesize {window}\\ \footnotesize{pixel} \\ \footnotesize{ranking}};
\node (n9) [alg]  {\footnotesize {window } \\ \footnotesize{kernel} \\\footnotesize{correlation}};
\node (n10) [alg]  {\footnotesize{spearman} \\\footnotesize{ranking}\\ \footnotesize{matrix}};
    \end{scope}
\node[draw,above=of n1] [circle] {1};
\node[draw,above=of n2,circle][circle]  {2};
\node[draw,above=of n3,circle] [circle] {3};
\node[draw,above=of n4,circle][circle]  {4};
\node[draw,above=of n5,circle] [circle] {5};
\node[draw,above=of n6,circle] [circle] {6};
\node[draw,above=of n7,circle] [circle] {7};
\node[draw,above=of n8,circle][circle]  {8};
\node[draw,above=of n9,circle] [circle] {9};
\node[draw,above=of n10] [circle] {10};
    \end{tikzpicture}

%% file: TypeB.tikz
\begin{tikzpicture}[scale=0.85,multilayer=2d,rotate=-90]
\begin{Layer}[layer=2]

\node at ( -2.3,-4)[rotate=90]{\footnotesize 4 X 4};
\node at ( 2.2,-4)[rotate=90]{\footnotesize 9 X 9};
\node at ( 6.2,-4)[rotate=90]{\footnotesize 13 X 13};

\node at ( -2.55,-0.7)[rotate=90]{\footnotesize 16n.};
\node at ( 2,-1.1)[rotate=90]{\footnotesize 81n.};
\node at ( 6.35,-0.6)[rotate=90]{\footnotesize 169.};

\end{Layer}

\Vertex[x=0.7,y=-0.5,layer=3,color=lightgray,opacity=1]{F4}
\Vertex[x=-0.2,y=-0.5,layer=3,color=lightgray,opacity=1]{F3}
\Vertex[x=-1.1,y=-0.5,layer=3,color=lightgray,opacity=1]{F2}
\Vertex[x=-2,y=-0.5,layer=3,color=lightgray,opacity=1]{F1}

\Vertex[x=2.6,y=-1.2,layer=3,color=lightgray,opacity=1]{F5}
\Vertex[x=3.5,y=-1.2,layer=3,color=lightgray,opacity=1]{F6}
\Vertex[x=4.4,y=-1.2,layer=3,color=lightgray,opacity=1]{F7}
\Vertex[x=5.3,y=-1.2,layer=3,color=lightgray,opacity=1]{F8}

\Vertex[x=7,y=-0.5,layer=3,color=lightgray,opacity=1]{F9}
\Vertex[x=7.9,y=-0.5,layer=3,color=lightgray,opacity=1]{F10}
\Vertex[x=8.8,y=-0.5,layer=3,color=lightgray,opacity=1]{F11}
\Vertex[x=9.7,y=-0.5,layer=3,color=lightgray,opacity=1]{F12}

\Vertex[x=9.7,y=1,layer=3,color=lightgray,opacity=1]{h12}
\Vertex[x=8.8,y=1,layer=3,color=lightgray,opacity=1]{h11}
\Vertex[x=7.9,y=1,layer=3,color=lightgray,opacity=1]{h10}
\Vertex[x=7,y=1,layer=3,color=lightgray,opacity=1]{h9}

\Vertex[x=5.7,y=1,layer=3,color=lightgray,opacity=1]{h8}
\Vertex[x=4.8,y=1,layer=3,color=lightgray,opacity=1]{h7}
\Vertex[x=3.9,y=1,layer=3,color=lightgray,opacity=1]{h6}
\Vertex[x=3,y=1,layer=3,color=lightgray,opacity=1]{h5}

\Vertex[x=0.8,y=1,layer=3,color=lightgray,opacity=1]{h4}
\Vertex[x=-0.2,y=1,layer=3,color=lightgray,opacity=1]{h3}
\Vertex[x=-1.1,y=1,layer=3,color=lightgray,opacity=1]{h2}
\Vertex[x=-2,y=1,layer=3,color=lightgray,opacity=1]{h1}

\Vertex[x=2.5,y=3,layer=2,color=lightgray,opacity=1]{o1}
\Vertex[x=3.5,y=3,layer=2,color=lightgray,opacity=1]{o2}
\Vertex[x=4.5,y=3,layer=2,color=lightgray,opacity=1]{o3}
\Vertex[x=5.5,y=3,layer=2,color=lightgray,opacity=1]{o4}
\Vertex[x=6.5,y=3,layer=2,color=lightgray,opacity=1]{o5}

\begin{Layer}[layer=3]

\draw[fill] (10.3,-0.5) circle [radius=0.1];
\draw[fill] (10.6,-0.5) circle [radius=0.07];
\draw[fill] (10.9,-0.5) circle [radius=0.05];

\draw[fill] (1.4,-0.5) circle [radius=0.1];
\draw[fill] (1.7,-0.5) circle [radius=0.07];
\draw[fill] (2.0,-0.5) circle [radius=0.05];

\draw[fill] (5.9,-1.2) circle [radius=0.1];
\draw[fill] (6.2,-1.2) circle [radius=0.07];
\draw[fill] (6.6,-1.2) circle [radius=0.05];

\draw[fill] (10.3,1) circle [radius=0.1];
\draw[fill] (10.6,1) circle [radius=0.07];
\draw[fill] (10.9,1) circle [radius=0.05];

\draw[fill] (1.4,1) circle [radius=0.1];
\draw[fill] (1.7,1) circle [radius=0.07];
\draw[fill] (2.0,1) circle [radius=0.05];

\draw[fill] (6.4,1) circle [radius=0.07];

\draw[fill] (7.2,3) circle [radius=0.1];
\draw[fill] (7.5,3) circle [radius=0.07];
\draw[fill] (7.8,3) circle [radius=0.05];

\node [rotate=90] at (9.7,-0.5) {\tiny$F_{12}$};
\node [rotate=90] at (8.8,-0.5) {\tiny$F_{11}$};
\node [rotate=90] at (7.9,-0.5) {\tiny$F_{10}$};
\node [rotate=90] at (7,-0.5) {\tiny$F_9$};

\node [rotate=90] at (5.3,-1.2) {\tiny$F_8$};
\node [rotate=90] at (4.4,-1.2) {\tiny$F_7$};
\node [rotate=90] at (3.5,-1.2) {\tiny$F_6$};
\node [rotate=90] at (2.6,-1.2) {\tiny$F_5$};

\node [rotate=90] at (0.7,-0.5) {\tiny$F_4$};
\node [rotate=90] at (-0.2,-0.5) {\tiny$F_3$};
\node [rotate=90] at (-1.1,-0.5) {\tiny$F_2$};
\node [rotate=90] at (-2,-0.5) {\tiny$F_1$};

\node [rotate=90] at (-2,1) {\tiny$H_1$};
\node [rotate=90] at (-1.1,1) {\tiny$H_2$};
\node [rotate=90] at (-0.2,1) {\tiny$H_3$};
\node [rotate=90] at (0.7,1) {\tiny$H_4$};

\node [rotate=90] at (3,1) {\tiny$H_5$};
\node [rotate=90] at (3.9,1) {\tiny$H_6$};
\node [rotate=90] at (4.8,1) {\tiny$H_7$};
\node [rotate=90] at (5.7,1) {\tiny$H_8$};

\node [rotate=90] at (7,1) {\tiny$H_9$};
\node [rotate=90] at (7.9,1) {\tiny$H_{10}$};
\node [rotate=90] at (8.8,1) {\tiny$H_{11}$};
\node [rotate=90] at (9.7,1) {\tiny$H_{12}$};

\node [rotate=90] at (2.5,3) {\tiny$O_1$};
\node [rotate=90] at (3.5,3) {\tiny$O_2$};
\node [rotate=90] at (4.5,3) {\tiny$O_3$};
\node [rotate=90] at (5.5,3) {\tiny$O_4$};
\node [rotate=90] at (6.5,3) {\tiny$O_5$};

\end{Layer}
\Edge[Math,distance=0.35,fontscale=0.5,Direct,lw=0.2,bend=0](F1)(h1)
\Edge[Math,distance=0.2,fontscale=0.5,Direct,lw=0.2,bend=0](F1)(h2)
\Edge[Math,distance=0.2,fontscale=0.5,Direct,lw=0.2,bend=0](F1)(h3)
\Edge[Math,distance=0.2,fontscale=0.5,Direct,lw=0.2,bend=0](F1)(h4)

\Edge[Math,distance=0.32,fontscale=0.5,Direct,lw=0.2,bend=0](F2)(h1)
\Edge[Math,distance=0.14,fontscale=0.5,Direct,lw=0.2,bend=0](F2)(h2)
\Edge[Math,distance=0.14,fontscale=0.5,Direct,lw=0.2,bend=0](F2)(h3)
\Edge[Math,distance=0.2,fontscale=0.5,Direct,lw=0.2,bend=0](F2)(h4)

\Edge[Math,distance=0.66,fontscale=0.5,Direct,lw=0.2,bend=0](F3)(h1)
\Edge[Math,distance=0.65,fontscale=0.5,Direct,lw=0.2,bend=0](F3)(h2)
\Edge[Math,distance=0.65,fontscale=0.5,Direct,lw=0.2,bend=0](F3)(h3)
\Edge[Math,distance=0.2,fontscale=0.5,Direct,lw=0.2,bend=0](F3)(h4)

\Edge[Math,distance=0.1,fontscale=0.5,Direct,lw=0.2,bend=0](F4)(h1)
\Edge[Math,distance=0.25,fontscale=0.5,Direct,lw=0.2,bend=0](F4)(h2)
\Edge[Math,distance=0.25,fontscale=0.5,Direct,lw=0.2,bend=0](F4)(h3)
\Edge[Math,distance=0.2,fontscale=0.5,Direct,lw=0.2,bend=0](F4)(h4)

\Edge[Math,distance=0.1,fontscale=0.5,Direct,lw=0.2,bend=0](F5)(h5)
\Edge[Math,distance=0.25,fontscale=0.5,Direct,lw=0.2,bend=0](F5)(h6)
\Edge[Math,distance=0.25,fontscale=0.5,Direct,lw=0.2,bend=0](F5)(h7)
\Edge[Math,distance=0.2,fontscale=0.5,Direct,lw=0.2,bend=0](F5)(h8)

\Edge[Math,distance=0.1,fontscale=0.5,Direct,lw=0.2,bend=0](F6)(h5)
\Edge[Math,distance=0.25,fontscale=0.5,Direct,lw=0.2,bend=0](F6)(h6)
\Edge[Math,distance=0.25,fontscale=0.5,Direct,lw=0.2,bend=0](F6)(h7)
\Edge[Math,distance=0.2,fontscale=0.5,Direct,lw=0.2,bend=0](F6)(h8)

\Edge[Math,distance=0.1,fontscale=0.5,Direct,lw=0.2,bend=0](F7)(h5)
\Edge[Math,distance=0.25,fontscale=0.5,Direct,lw=0.2,bend=0](F7)(h6)
\Edge[Math,distance=0.25,fontscale=0.5,Direct,lw=0.2,bend=0](F7)(h7)
\Edge[Math,distance=0.2,fontscale=0.5,Direct,lw=0.2,bend=0](F7)(h8)

\Edge[Math,distance=0.1,fontscale=0.5,Direct,lw=0.2,bend=0](F8)(h5)
\Edge[Math,distance=0.25,fontscale=0.5,Direct,lw=0.2,bend=0](F8)(h6)
\Edge[Math,distance=0.25,fontscale=0.5,Direct,lw=0.2,bend=0](F8)(h7)
\Edge[Math,distance=0.2,fontscale=0.5,Direct,lw=0.2,bend=0](F8)(h8)

\Edge[Math,distance=0.1,fontscale=0.5,Direct,lw=0.2,bend=0](F9)(h9)
\Edge[Math,distance=0.25,fontscale=0.5,Direct,lw=0.2,bend=0](F9)(h10)
\Edge[Math,distance=0.25,fontscale=0.5,Direct,lw=0.2,bend=0](F9)(h11)
\Edge[Math,distance=0.2,fontscale=0.5,Direct,lw=0.2,bend=0](F9)(h12)

\Edge[Math,distance=0.1,fontscale=0.5,Direct,lw=0.2,bend=0](F10)(h9)
\Edge[Math,distance=0.25,fontscale=0.5,Direct,lw=0.2,bend=0](F10)(h10)
\Edge[Math,distance=0.25,fontscale=0.5,Direct,lw=0.2,bend=0](F10)(h11)
\Edge[Math,distance=0.2,fontscale=0.5,Direct,lw=0.2,bend=0](F10)(h12)

\Edge[Math,distance=0.1,fontscale=0.5,Direct,lw=0.2,bend=0](F11)(h9)
\Edge[Math,distance=0.25,fontscale=0.5,Direct,lw=0.2,bend=0](F11)(h10)
\Edge[Math,distance=0.25,fontscale=0.5,Direct,lw=0.2,bend=0](F11)(h11)
\Edge[Math,distance=0.2,fontscale=0.5,Direct,lw=0.2,bend=0](F11)(h12)

\Edge[Math,distance=0.1,fontscale=0.5,Direct,lw=0.2,bend=0](F12)(h9)
\Edge[Math,distance=0.25,fontscale=0.5,Direct,lw=0.2,bend=0](F12)(h10)
\Edge[Math,distance=0.25,fontscale=0.5,Direct,lw=0.2,bend=0](F12)(h11)
\Edge[Math,distance=0.2,fontscale=0.5,Direct,lw=0.2,bend=0](F12)(h12)

\Edge[Math,distance=0.21,fontscale=0.5,lw=0.2,Direct,bend=0](h1)(o1)
\Edge[Math,distance=0.25,fontscale=0.5,lw=0.2,Direct,bend=0](h1)(o2)
\Edge[Math,distance=0.25,fontscale=0.5,lw=0.2,Direct,bend=0](h1)(o3)
\Edge[Math,distance=0.25,fontscale=0.5,lw=0.2,Direct,bend=0](h1)(o4)
\Edge[Math,distance=0.25,fontscale=0.5,lw=0.2,Direct,bend=0](h1)(o5)

\Edge[Math,distance=0.21,fontscale=0.5,lw=0.2,Direct,bend=0](h2)(o1)
\Edge[Math,distance=0.25,fontscale=0.5,lw=0.2,Direct,bend=0](h2)(o2)
\Edge[Math,distance=0.25,fontscale=0.5,lw=0.2,Direct,bend=0](h2)(o3)
\Edge[Math,distance=0.25,fontscale=0.5,lw=0.2,Direct,bend=0](h2)(o4)
\Edge[Math,distance=0.25,fontscale=0.5,lw=0.2,Direct,bend=0](h2)(o5)

\Edge[Math,distance=0.21,fontscale=0.5,lw=0.2,Direct,bend=0](h3)(o1)
\Edge[Math,distance=0.25,fontscale=0.5,lw=0.2,Direct,bend=0](h3)(o2)
\Edge[Math,distance=0.25,fontscale=0.5,lw=0.2,Direct,bend=0](h3)(o3)
\Edge[Math,distance=0.25,fontscale=0.5,lw=0.2,Direct,bend=0](h3)(o4)
\Edge[Math,distance=0.25,fontscale=0.5,lw=0.2,Direct,bend=0](h3)(o5)

\Edge[Math,distance=0.21,fontscale=0.5,lw=0.2,Direct,bend=0](h4)(o1)
\Edge[Math,distance=0.25,fontscale=0.5,lw=0.2,Direct,bend=0](h4)(o2)
\Edge[Math,distance=0.25,fontscale=0.5,lw=0.2,Direct,bend=0](h4)(o3)
\Edge[Math,distance=0.25,fontscale=0.5,lw=0.2,Direct,bend=0](h4)(o4)
\Edge[Math,distance=0.25,fontscale=0.5,lw=0.2,Direct,bend=0](h4)(o5)

\Edge[Math,distance=0.21,fontscale=0.5,lw=0.2,Direct,bend=0](h5)(o1)
\Edge[Math,distance=0.25,fontscale=0.5,lw=0.2,Direct,bend=0](h5)(o2)
\Edge[Math,distance=0.25,fontscale=0.5,lw=0.2,Direct,bend=0](h5)(o3)
\Edge[Math,distance=0.25,fontscale=0.5,lw=0.2,Direct,bend=0](h5)(o4)
\Edge[Math,distance=0.25,fontscale=0.5,lw=0.2,Direct,bend=0](h5)(o5)

\Edge[Math,distance=0.21,fontscale=0.5,lw=0.2,Direct,bend=0](h6)(o1)
\Edge[Math,distance=0.25,fontscale=0.5,lw=0.2,Direct,bend=0](h6)(o2)
\Edge[Math,distance=0.25,fontscale=0.5,lw=0.2,Direct,bend=0](h6)(o3)
\Edge[Math,distance=0.25,fontscale=0.5,lw=0.2,Direct,bend=0](h6)(o4)
\Edge[Math,distance=0.25,fontscale=0.5,lw=0.2,Direct,bend=0](h6)(o5)

\Edge[Math,distance=0.21,fontscale=0.5,lw=0.2,Direct,bend=0](h7)(o1)
\Edge[Math,distance=0.25,fontscale=0.5,lw=0.2,Direct,bend=0](h7)(o2)
\Edge[Math,distance=0.25,fontscale=0.5,lw=0.2,Direct,bend=0](h7)(o3)
\Edge[Math,distance=0.25,fontscale=0.5,lw=0.2,Direct,bend=0](h7)(o4)
\Edge[Math,distance=0.25,fontscale=0.5,lw=0.2,Direct,bend=0](h7)(o5)

\Edge[Math,distance=0.21,fontscale=0.5,lw=0.2,Direct,bend=0](h8)(o1)
\Edge[Math,distance=0.25,fontscale=0.5,lw=0.2,Direct,bend=0](h8)(o2)
\Edge[Math,distance=0.25,fontscale=0.5,lw=0.2,Direct,bend=0](h8)(o3)
\Edge[Math,distance=0.25,fontscale=0.5,lw=0.2,Direct,bend=0](h8)(o4)
\Edge[Math,distance=0.25,fontscale=0.5,lw=0.2,Direct,bend=0](h8)(o5)

\Edge[Math,distance=0.21,fontscale=0.5,lw=0.2,Direct,bend=0](h9)(o1)
\Edge[Math,distance=0.25,fontscale=0.5,lw=0.2,Direct,bend=0](h9)(o2)
\Edge[Math,distance=0.25,fontscale=0.5,lw=0.2,Direct,bend=0](h9)(o3)
\Edge[Math,distance=0.25,fontscale=0.5,lw=0.2,Direct,bend=0](h9)(o4)
\Edge[Math,distance=0.25,fontscale=0.5,lw=0.2,Direct,bend=0](h9)(o5)

\Edge[Math,distance=0.21,fontscale=0.5,lw=0.2,Direct,bend=0](h10)(o1)
\Edge[Math,distance=0.25,fontscale=0.5,lw=0.2,Direct,bend=0](h10)(o2)
\Edge[Math,distance=0.25,fontscale=0.5,lw=0.2,Direct,bend=0](h10)(o3)
\Edge[Math,distance=0.25,fontscale=0.5,lw=0.2,Direct,bend=0](h10)(o4)
\Edge[Math,distance=0.25,fontscale=0.5,lw=0.2,Direct,bend=0](h10)(o5)

\Edge[Math,distance=0.21,fontscale=0.5,lw=0.2,Direct,bend=0](h11)(o1)
\Edge[Math,distance=0.25,fontscale=0.5,lw=0.2,Direct,bend=0](h11)(o2)
\Edge[Math,distance=0.25,fontscale=0.5,lw=0.2,Direct,bend=0](h11)(o3)
\Edge[Math,distance=0.25,fontscale=0.5,lw=0.2,Direct,bend=0](h11)(o4)
\Edge[Math,distance=0.25,fontscale=0.5,lw=0.2,Direct,bend=0](h11)(o5)

\Edge[Math,distance=0.21,fontscale=0.5,lw=0.2,Direct,bend=0](h12)(o1)
\Edge[Math,distance=0.25,fontscale=0.5,lw=0.2,Direct,bend=0](h12)(o2)
\Edge[Math,distance=0.25,fontscale=0.5,lw=0.2,Direct,bend=0](h12)(o3)
\Edge[Math,distance=0.25,fontscale=0.5,lw=0.2,Direct,bend=0](h12)(o4)
\Edge[Math,distance=0.25,fontscale=0.5,lw=0.2,Direct,bend=0](h12)(o5)

\end{tikzpicture}

%% file: kernel3.tikz
\begin{tikzpicture}[scale=.5]

\newcommand\setrow[5]{
  \setcounter{col}{1}
  \foreach \n in {#1, #2, #3, #4, #5} {
    \edef\x{\value{col} - 0.5}
    \edef\y{5.5 - \value{row}}
    \node[anchor=center] at (\x, \y) {\n};
    \stepcounter{col}
  }
  \stepcounter{row}
}

  \begin{scope}
    \draw (0, 0) grid (5, 5);
    \draw[very thick, scale=1] (0, 0) grid (5, 5);

\foreach \y in {0,1,...,5}
{\draw (0,\y) -- (5.5,\y);}

\foreach \x in {0,1,...,5}
{\draw (\x,0) -- (\x,5.5);}

    \setcounter{row}{1}
    \setrow {}{}{}  {} {}
    \setrow {}{}{} {} {}
    \setrow {}{}{}  {}{}
    \setrow {}{}{}  {} {}
    \setrow {}{}{}  {} {}

  \end{scope}
  \end{tikzpicture}

%% file: kernel2.tikz

\newcommand\setrow[4]{
  \setcounter{col}{1}
  \foreach \n in {#1, #2, #3, #4} {
    \edef\x{\value{col} - 0.5}
    \edef\y{4.5 - \value{row}}
    \node[anchor=center] at (\x, \y) {\n};
    \stepcounter{col}
  }
  \stepcounter{row}
}

\begin{tikzpicture}[scale=.5,rotate=90]

  \begin{scope}

    \draw[very thick, scale=1] (0, 0) grid (4, 4);
    
    \foreach \y in {0,1,...,4}
{\draw (0,\y) -- (4.5,\y);}
\foreach \x in {0,1,...,4}
{\draw (\x,0) -- (\x,4.5);}

    \setcounter{row}{1}
    \setrow {}{}{}{} 
    \setrow {}{}{}{}  
    \setrow {}{}{}{} 
    \setrow {}{}{}{}

  \end{scope}
  
\end{tikzpicture}

%% file: kernel1.tikz

\newcommand\setrow[3]{
  \setcounter{col}{1}
  \foreach \n in {#1, #2, #3} {
    \edef\x{\value{col} - 0.5}
    \edef\y{3.5 - \value{row}}
    \node[anchor=center] at (\x, \y) {\n};
    \stepcounter{col}
  }
  \stepcounter{row}
}

\begin{tikzpicture}[scale=.5,rotate=90]

  \begin{scope}

    \draw[very thick, scale=1] (0, 0) grid (3, 3);
    \setcounter{row}{1}
    \setrow {}{}{}   
    \setrow {}{}{}  
    \setrow {}{}{} 

  \end{scope}
  
\end{tikzpicture}

%% file: spearmangrid1.tikz
\begin{tikzpicture}[scale=.5]


\newcommand\setrow[4]{
  \setcounter{col}{1}
  \foreach \n in {#1, #2, #3, #4} {
    \edef\x{\value{col} - 0.5}
    \edef\y{4.5 - \value{row}}
    \node[anchor=center] at (\x, \y) {\n};
    \stepcounter{col}
  }
  \stepcounter{row}
}

  \begin{scope}
    \draw (0, 0) grid (4, 4);
     \draw[very thick, scale=2] (0, 0) grid (2, 2);
     \draw[very thick, scale=1] (0, 0) grid (1, 1);
    \setcounter{row}{1}
    \setrow {}{}{}{} 
    \setrow {}{}{}{} 
    \setrow {}{}{}{}
    \setrow {}{}{}{}

  \end{scope}
  \end{tikzpicture}

%% file: spearmangrid2.tikz
\begin{tikzpicture}[scale=.5]


\newcommand\setrow[6]{
  \setcounter{col}{1}
  \foreach \n in {#1, #2, #3, #4, #5, #6} {
    \edef\x{\value{col} - 0.5}
    \edef\y{6.5 - \value{row}}
    \node[anchor=center] at (\x, \y) {\n};
    \stepcounter{col}
  }
  \stepcounter{row}
}

  \begin{scope}
    \draw (0, 0) grid (6, 6);
     \draw[very thick, scale=2] (0, 0) grid (3, 3);
     \draw[very thick, scale=1] (0, 0) grid (1, 1);
    \setcounter{row}{1}
    \setrow {}{}{}{} {}  {}
    \setrow {}{}{}{}{}{}
    \setrow {}{}{}{}{}  {}
    \setrow {}{}{}{} {}  {}
     \setrow {}{}{}{}{}  {}
    \setrow {}{}{}{}{}  {}

  \end{scope}
  \end{tikzpicture}

%% file: spearmangrid3.tikz
\begin{tikzpicture}[scale=.5]

\newcommand\setrow[8]{
  \setcounter{col}{1}
  \foreach \n in {#1, #2, #3, #4, #5, #6, #7, #8} {
    \edef\x{\value{col} - 0.5}
    \edef\y{8.5 - \value{row}}
    \node[anchor=center] at (\x, \y) {\n};
    \stepcounter{col}
  }
  \stepcounter{row}
}

  \begin{scope}
    \draw (0, 0) grid (8, 8);
     \draw[very thick, scale=2] (0, 0) grid (4, 4);
     \draw[very thick, scale=1] (0, 0) grid (1, 1);

\foreach \y in {0,1,...,8}
{\draw (0,\y) -- (8.5,\y);}

\foreach \x in {0,1,...,8}
{\draw (\x,0) -- (\x,8.5);}

    \setcounter{row}{1}
    \setrow {}{}{}  {} {}{}{}  {} 
    \setrow {}{}{} {} {}{}{}  {} 
    \setrow {}{}{}  {}{}{}{}  {} 
    \setrow { }{}{}  {} {}{}{}  {} 
    \setrow {}{}{}  {} {}{}{}  {} 
    \setrow {}{}{} {} {}{}{}  {} 
    \setrow {}{}{}  {}{}{}{}  {}
    \setrow { }{}{}  {} {}{}{}  {} 

  \end{scope}
  \end{tikzpicture}

%% file: maxpoolinggrid1.tikz

\newcommand\setrow[2]{
  \setcounter{col}{1}
  \foreach \n in {#1, #2} {
    \edef\x{\value{col} - 0.5}
    \edef\y{2.5 - \value{row}}
    \node[anchor=center] at (\x, \y) {\n};
    \stepcounter{col}
  }
  \stepcounter{row}
}

\begin{tikzpicture}[scale=.5]

  \begin{scope}
    \draw (0, 0) grid (2, 2);
    \setcounter{row}{1}
    \setrow {}{}
    \setrow {}{}

  \end{scope}
  \end{tikzpicture}

%% file: maxpoolinggrid2.tikz

\newcommand\setrow[3]{
  \setcounter{col}{1}
  \foreach \n in {#1, #2, #3} {
    \edef\x{\value{col} - 0.5}
    \edef\y{3.5 - \value{row}}
    \node[anchor=center] at (\x, \y) {\n};
    \stepcounter{col}
  }
  \stepcounter{row}
}

\begin{tikzpicture}[scale=.5]

  \begin{scope}
    \draw (0, 0) grid (3, 3);
    \setcounter{row}{1}
    \setrow {}{}{}
    \setrow {}{}{}
 \setrow {}{}{}

  \end{scope}
  \end{tikzpicture}

%% file: maxpoolinggrid3.tikz

\newcommand\setrow[5]{
  \setcounter{col}{1}
  \foreach \n in {#1, #2, #3, #4, #5} {
    \edef\x{\value{col} - 0.5}
    \edef\y{5.5 - \value{row}}
    \node[anchor=center] at (\x, \y) {\n};
    \stepcounter{col}
  }
  \stepcounter{row}
}

\begin{tikzpicture}[scale=.5]

  \begin{scope}
    \draw (0, 0) grid (5, 5);
    \setcounter{row}{1}
    \setrow {}{}{}{}{}
    \setrow {}{}{}{}{}
   \setrow {}{}{}{}{}
    \setrow {}{}{}{}{}
 \setrow {}{}{}{}{}
  \end{scope}
  \end{tikzpicture}

%% file: main.bbl
\begin{thebibliography}{10}
\providecommand{\url}[1]{#1}
\csname url@samestyle\endcsname
\providecommand{\newblock}{\relax}
\providecommand{\bibinfo}[2]{#2}
\providecommand{\BIBentrySTDinterwordspacing}{\spaceskip=0pt\relax}
\providecommand{\BIBentryALTinterwordstretchfactor}{4}
\providecommand{\BIBentryALTinterwordspacing}{\spaceskip=\fontdimen2\font plus
\BIBentryALTinterwordstretchfactor\fontdimen3\font minus
  \fontdimen4\font\relax}
\providecommand{\BIBforeignlanguage}[2]{{%
\expandafter\ifx\csname l@#1\endcsname\relax
\typeout{** WARNING: IEEEtran.bst: No hyphenation pattern has been}%
\typeout{** loaded for the language `#1'. Using the pattern for}%
\typeout{** the default language instead.}%
\else
\language=\csname l@#1\endcsname
\fi
#2}}
\providecommand{\BIBdecl}{\relax}
\BIBdecl

\bibitem{PLbook}
J.~Frnkranz and E.~H{\"u}llermeier, \emph{Preference Learning}, 1st~ed.\hskip
  1em plus 0.5em minus 0.4em\relax Berlin, Heidelberg: Springer-Verlag, 2010.

\bibitem{prafman}
R.~Brafman and C.~Domshlak, ``Preference handling - an introductory tutorial,''
  pp. 58--86, 2009.

\bibitem{adom}
G.~Adomavicius and A.~Tuzhilin., ``Toward the next generation of recommender
  systems: a survey of the state-of-the-art and possible extensions,'' pp.
  734--749, 2005.

\bibitem{recommender}
M.~Montaner and B.~López, ``A taxonomy of recommender agents on the
  internet.'' pp. 285--330, 2003.

\bibitem{aio}
F.~Aiolli, ``A preference model for structured supervised learning tasks,'' pp.
  557--560, 2005.

\bibitem{crammer}
K.~Crammer and Y.~Singer, ``Pranking with ranking,'' pp. 641--647, 2002.

\bibitem{9714196}
Q.~Ni, J.~Guo, W.~Wu, and H.~Wang, ``Influence-based community partition with
  sandwich method for social networks,'' \emph{IEEE Transactions on
  Computational Social Systems}, pp. 1--12, 2022.

\bibitem{10.1093/comjnl/bxaa168}
H.~Wang, Q.~Gao, H.~Li, H.~Wang, L.~Yan, and G.~Liu, ``{A Structural
  Evolution-Based Anomaly Detection Method for Generalized Evolving Social
  Networks},'' \emph{The Computer Journal}, vol.~65, no.~5, pp. 1189--1199, 12
  2020.

\bibitem{9750402}
C.-Q. Huang, F.~Jiang, Q.-H. Huang, X.-Z. Wang, Z.-M. Han, and W.-Y. Huang,
  ``Dual-graph attention convolution network for 3-d point cloud
  classification,'' \emph{IEEE Transactions on Neural Networks and Learning
  Systems}, pp. 1--13, 2022.

\bibitem{10004751}
H.~Wang, Z.~Cui, R.~Liu, L.~Fang, and Y.~Sha, ``A multi-type transferable
  method for missing link prediction in heterogeneous social networks,''
  \emph{IEEE Transactions on Knowledge and Data Engineering}, pp. 1--13, 2022.

\bibitem{GUO202283}
F.~Guo, W.~Zhou, Q.~Lu, and C.~Zhang, ``Path extension similarity link
  prediction method based on matrix algebra in directed networks,''
  \emph{Computer Communications}, vol. 187, pp. 83--92, 2022.

\bibitem{electronics11193022}
X.~Qin, Z.~Liu, Y.~Liu, S.~Liu, B.~Yang, L.~Yin, M.~Liu, and W.~Zheng, ``User
  ocean personality model construction method using a bp neural network,''
  \emph{Electronics}, vol.~11, no.~19, 2022.

\bibitem{9983500}
L.~Liu, S.~Zhang, L.~Zhang, G.~Pan, and J.~Yu, ``Multi-uuv maneuvering
  counter-game for dynamic target scenario based on fractional-order recurrent
  neural network,'' \emph{IEEE Transactions on Cybernetics}, pp. 1--14, 2022.

\bibitem{taxonamy}
Y.~Zhou, Y.~Liu, J.~Yang, X.~He, and L.~Liu, ``A taxonomy of label ranking
  algorithms,'' \emph{JCP}, vol.~9, pp. 557--565, 2014.

\bibitem{Furnkranz1}
J.~Furnkranz and E.~H{\"u}llermeier, ``Pairwise preference learning and ranking
  in machine learning,'' pp. 145--156, 2003.

\bibitem{Furnkranz2}
J.~F{\"u}rnkranz and E.~H{\"u}llermeier, ``Preference learning,'' 2010.

\bibitem{constraint}
S.~Har-Peled, D.~Roth, and D.~Zimak, ``Constraint classification: A new
  approach to multiclass classification,'' 2002.

\bibitem{Hullermeier}
E.~H{\"u}llermeier, J.~Furnkranz, W.~Cheng, and K.~Brinker, ``Label ranking by
  learning pairwise preferences,'' pp. 1897--1916, 2008.

\bibitem{dt1}
J.~Furnkranz and E.~H{\"u}llermeier, ``Decision tree modeling for ranking
  data,'' pp. 83--106, 2011.

\bibitem{plackett}
W.~Cheng and E.~H{\"u}llermeier, ``Instance-based label ranking using the
  mallows model,'' pp. 143--157, 2008.

\bibitem{gaussian}
G.~Mihajlo, D.~Nemanja, and V.~Slobodan, ``Learning from pairwise preference
  data using gaussian mixture model,'' 2014.

\bibitem{Burges}
T.~S. Chris~Burges, ``Learning to rank using gradient descent,'' pp. 58--86,
  2005.

\bibitem{Ribeiro}
G.~Ribeiro, W.~Duivesteijn, C.~Soares, and A.~Knobbe, ``Multilayer perceptron
  for label ranking,'' in \emph{Proceedings of the 22nd International
  Conference on Artificial Neural Networks and Machine Learning - Volume Part
  II}, ser. ICANN’12.\hskip 1em plus 0.5em minus 0.4em\relax Springer, 2012,
  p. 25–32.

\bibitem{Freund}
Y.~Freund, R.~Iyer, R.~E. Schapire, and Y.~Singer, ``An efficient boosting
  algorithm for combining preferences,'' \emph{J. Mach. Learn. Res.}, vol.~4,
  no. null, p. 933–969, Dec. 2003.

\bibitem{svore}
\BIBentryALTinterwordspacing
Q.~Wu, C.~J. Burges, K.~M. Svore, and J.~Gao, ``Adapting boosting for
  information retrieval measures,'' vol.~13, no.~3, p. 254–270, Jun. 2010.
  [Online]. Available: \url{https://doi.org/10.1007/s10791-009-9112-1}
\BIBentrySTDinterwordspacing

\bibitem{deeppairwise}
Y.~{Jian}, J.~{Xiao}, Y.~{Cao}, A.~{Khan}, and J.~{Zhu}, ``Deep pairwise
  ranking with multi-label information for cross-modal retrieval,'' in
  \emph{2019 IEEE International Conference on Multimedia and Expo (ICME)},
  2019, pp. 1810--1815.

\bibitem{Deeplabelraning}
J.~Li, W.~Y. Ng~Wing, T.~Xing, S.~Kwong, and H.~Wang, ``Weighted multi-deep
  ranking supervised hashing for efficient image retrieval,''
  \emph{International Journal of Machine Learning and Cybernetics}, vol.~11,
  no.~4, pp. 883--897, 2020.

\bibitem{labelclass}
Z.~Ji, B.~Cui, H.~Li, Y.-G. Jiang, T.~Xiang, T.~Hospedales, and Y.~Fu, ``Deep
  ranking for image zero-shot multi-label classification,'' \emph{IEEE
  transactions on image processing : a publication of the IEEE Signal
  Processing Society}, May 14 2020.

\bibitem{Cherian2021ClassificationOR}
A.~K. Cherian and E.~Poovammal, ``Classification of remote sensing images using
  cnn,'' \emph{IOP Conference Series: Materials Science and Engineering}, vol.
  1130, 2021.

\bibitem{singh}
A.~Robert~Singh and S.~Athisayamani, ``Survival prediction based on brain tumor
  classification using convolutional neural network with channel preference,''
  in \emph{Data Engineering and Intelligent Computing}, V.~Bhateja,
  L.~Khin~Wee, J.~C.-W. Lin, S.~C. Satapathy, and T.~M. Rajesh, Eds.\hskip 1em
  plus 0.5em minus 0.4em\relax Singapore: Springer Nature Singapore, 2022, pp.
  259--269.

\bibitem{9137336}
Z.~Lv, L.~Qiao, J.~Li, and H.~Song, ``Deep-learning-enabled security issues in
  the internet of things,'' \emph{IEEE Internet of Things Journal}, vol.~8,
  no.~12, pp. 9531--9538, 2021.

\bibitem{10.1145/3468506}
Z.~Lv, Z.~Yu, S.~Xie, and A.~Alamri, ``Deep learning-based smart predictive
  evaluation for interactive multimedia-enabled smart healthcare,'' \emph{ACM
  Trans. Multimedia Comput. Commun. Appl.}, vol.~18, no.~1s, jan 2022.

\bibitem{9999261}
J.~Xu, S.~Pan, P.~Z.~H. Sun, S.~Hyeong~Park, and K.~Guo,
  ``Human-factors-in-driving-loop: Driver identification and verification via a
  deep learning approach using psychological behavioral data,'' \emph{IEEE
  Transactions on Intelligent Transportation Systems}, vol.~24, no.~3, pp.
  3383--3394, 2023.

\bibitem{https://doi.org/10.1029/2022WR033241}
C.~Zhan, Z.~Dai, M.~R. Soltanian, and F.~P.~J. de~Barros, ``Data-worth analysis
  for heterogeneous subsurface structure identification with a stochastic deep
  learning framework,'' \emph{Water Resources Research}, vol.~58, no.~11, p.
  e2022WR033241, 2022.

\bibitem{rs13224604}
\BIBentryALTinterwordspacing
S.~Pare, H.~Mittal, M.~Sajid, J.~C. Bansal, A.~Saxena, T.~Jan, W.~Pedrycz, and
  M.~Prasad, ``Remote sensing imagery segmentation: A hybrid approach,''
  \emph{Remote Sensing}, vol.~13, no.~22, 2021. [Online]. Available:
  \url{https://www.mdpi.com/2072-4292/13/22/4604}
\BIBentrySTDinterwordspacing

\bibitem{moraga}
C.~Moraga and R.~Heider, ``"{N}ew lamps for old!" (generalized multiple-valued
  neurons),'' in \emph{Proceedings 1999 29th IEEE International Symposium on
  Multiple-Valued Logic (Cat. No. 99CB36329)}.\hskip 1em plus 0.5em minus
  0.4em\relax IEEE, 1999, pp. 36--41.

\bibitem{mvn1}
I.~Aizenberg, N.~Aizenberg, and J.~P. Vandewalle, \emph{Multi-Valued and
  Universal Binary Neurons: Theory, Learning and Applications}.\hskip 1em plus
  0.5em minus 0.4em\relax Norwell, MA, USA: Kluwer Academic Publishers, 2000.

\bibitem{8941002}
W.~Zhou, Y.~Lv, J.~Lei, and L.~Yu, ``Global and local-contrast guides
  content-aware fusion for rgb-d saliency prediction,'' \emph{IEEE Transactions
  on Systems, Man, and Cybernetics: Systems}, vol.~51, no.~6, pp. 3641--3649,
  2021.

\bibitem{10018569}
B.~Xie, S.~Li, M.~Li, C.~H. Liu, G.~Huang, and G.~Wang, ``Sepico:
  Semantic-guided pixel contrast for domain adaptive semantic segmentation,''
  \emph{IEEE Transactions on Pattern Analysis and Machine Intelligence}, pp.
  1--17, 2023.

\bibitem{https://doi.org/10.1049/cit2.12174}
X.~Zhang, D.~Huang, H.~Li, Y.~Zhang, Y.~Xia, and J.~Liu, ``Self-training
  maximum classifier discrepancy for eeg emotion recognition,'' \emph{CAAI
  Transactions on Intelligence Technology}, vol. n/a, no. n/a.

\bibitem{Jiang2020RobustVS}
F.~Jiang, B.~Kong, J.~Li, K.~Dashtipour, and M.~Gogate, ``Robust visual
  saliency optimization based on bidirectional markov chains,'' \emph{Cognitive
  Computation}, pp. 1--12, 2020.

\bibitem{Gupta2020SalientOD}
A.~K. Gupta, A.~Seal, M.~Prasad, and P.~Khanna, ``Salient object detection
  techniques in computer vision—a survey,'' \emph{Entropy}, vol.~22, 2020.

\bibitem{Lei2021SportsID}
H.~Lei, T.~Lei, and T.~Yue-nian, ``Sports image detection based on particle
  swarm optimization algorithm,'' \emph{Microprocess. Microsystems}, vol.~80,
  p. 103345, 2021.

\bibitem{ZHANG2022109766}
K.~Zhang, Z.~Wang, G.~Chen, L.~Zhang, Y.~Yang, C.~Yao, J.~Wang, and J.~Yao,
  ``Training effective deep reinforcement learning agents for real-time
  life-cycle production optimization,'' \emph{Journal of Petroleum Science and
  Engineering}, vol. 208, p. 109766, 2022.

\bibitem{app13053082}
M.~Liu, Q.~Gu, B.~Yang, Z.~Yin, S.~Liu, L.~Yin, and W.~Zheng, ``Kinematics
  model optimization algorithm for six degrees of freedom parallel platform,''
  \emph{Applied Sciences}, vol.~13, no.~5, 2023.

\bibitem{9350239}
G.~Zhou, R.~Zhang, and S.~Huang, ``Generalized buffering algorithm,''
  \emph{IEEE Access}, vol.~9, pp. 27\,140--27\,157, 2021.

\bibitem{Zhang2022SportsAR}
R.~Zhang, ``Sports action recognition based on particle swarm optimization
  neural networks,'' \emph{Wireless Communications and Mobile Computing}, 2022.

\bibitem{videofile}
\BIBentryALTinterwordspacing
A.~Elgharabawy, ``Preference neural network convergence performance,'' Video
  file, 2020. [Online]. Available:
  \url{https://drive.google.com/drive/folders/1yxuqYoQ3Kiuch-2sLeVe2ocMj12QVsRM?usp=sharing}
\BIBentrySTDinterwordspacing

\bibitem{sss}
G.~Bologna, ``Rule extraction from a multilayer perceptron with staircase
  activation functions,'' 2000.

\bibitem{kendall}
M.~Kendall, ``Rank correlation methods,'' 1948.

\bibitem{spears}
\BIBentryALTinterwordspacing
C.~Spearman, ``The proof and measurement of association between two things,''
  \emph{The American Journal of Psychology}, vol.~15, no.~1, pp. 72--101, 1904.
  [Online]. Available: \url{http://www.jstor.org/stable/1412159}
\BIBentrySTDinterwordspacing

\bibitem{Cheng}
W.~Cheng, J.~H\"{u}hn, and E.~H\"{u}llermeier, ``Decision tree and
  instance-based learning for label ranking,'' in \emph{Proceedings of the 26th
  Annual International Conference on Machine Learning}, ser. ICML ’09.\hskip
  1em plus 0.5em minus 0.4em\relax ACM, 2009, p. 161–168.

\bibitem{mnist}
\BIBentryALTinterwordspacing
Y.~LeCun and C.~Cortes, ``{MNIST} handwritten digit database,'' 2010. [Online].
  Available: \url{http://yann.lecun.com/exdb/mnist/}
\BIBentrySTDinterwordspacing

\bibitem{cfar}
A.~Krizhevsky, ``Learning multiple layers of features from tiny images,'' Tech.
  Rep., 2009.

\bibitem{mnistfashion}
\BIBentryALTinterwordspacing
H.~Xiao, K.~Rasul, and R.~Vollgraf, ``Fashion-mnist: a novel image dataset for
  benchmarking machine learning algorithms,'' 2017. [Online]. Available:
  \url{http://arxiv.org/abs/1708.07747}
\BIBentrySTDinterwordspacing

\bibitem{newdataset}
C.~R. de~S$\acute{a}$ and W.~Duivesteijn, ``Discovering a taste for the
  unusual: exceptional models for preference mining,'' pp. 1775--1807, 2018.

\bibitem{dataset1}
\BIBentryALTinterwordspacing
R.~Cláudio. (2018) algae dataset. [Online]. Available:
  \url{http://dx.doi.org/10.17632/3mv94c8jpc.2}
\BIBentrySTDinterwordspacing

\bibitem{Grbovic}
M.~Grbovic, N.~Djuric, S.~Guo, and S.~Vucetic, ``Supervised clustering of label
  ranking data using label preference information,'' \emph{Machine Learning},
  vol.~93, no. 2-3, pp. 191--225, 2013.

\bibitem{resnet}
K.~{He}, X.~{Zhang}, S.~{Ren}, and J.~{Sun}, ``Deep residual learning for image
  recognition,'' in \emph{2016 IEEE Conference on Computer Vision and Pattern
  Recognition (CVPR)}, 2016, pp. 770--778.

\bibitem{wrn}
S.~Zagoruyko and N.~Komodakis, ``Wide residual networks,'' \emph{ArXiv}, vol.
  abs/1605.07146, 2016.

\bibitem{dens}
G.~{Huang}, Z.~{Liu}, L.~{Van Der Maaten}, and K.~Q. {Weinberger}, ``Densely
  connected convolutional networks,'' in \emph{2017 IEEE Conference on Computer
  Vision and Pattern Recognition (CVPR)}, 2017, pp. 2261--2269.

\bibitem{Tan2021EfficientNetV2SM}
M.~Tan and Q.~V. Le, ``Efficientnetv2: Smaller models and faster training,''
  \emph{ArXiv}, vol. abs/2104.00298, 2021.

\bibitem{mlcoder}
T.~Ridnik, G.~Sharir, A.~Ben-Cohen, E.~Ben-Baruch, and A.~Noy, ``Ml-decoder:
  Scalable and versatile classification head,'' 2021.

\bibitem{Wu2021CvTIC}
H.~Wu, B.~Xiao, N.~C.~F. Codella, M.~Liu, X.~Dai, L.~Yuan, and L.~Zhang, ``Cvt:
  Introducing convolutions to vision transformers,'' \emph{2021 IEEE/CVF
  International Conference on Computer Vision (ICCV)}, pp. 22--31, 2021.

\bibitem{Zhang2018LSTMAI}
K.~Zhang, ``Lstm: An image classification model based on fashion-mnist
  dataset,'' 2018.

\bibitem{Tanveer2021FineTuningDF}
M.~Tanveer, M.~U.~K. Khan, and C.~M. Kyung, ``Fine-tuning darts for image
  classification,'' \emph{2020 25th International Conference on Pattern
  Recognition (ICPR)}, pp. 4789--4796, 2021.

\bibitem{forest}
C.~R. de~S{\'a}, C.~Soares, A.~Knobbe, and P.~Cortez, ``Label ranking
  forests,'' \emph{Expert Syst. J. Knowl. Eng.}, vol.~34, 2017.

\bibitem{pnngithub}
\BIBentryALTinterwordspacing
A.~Elgharabawy, ``Preference neural network source code,'' Python Code,
  Mathematica code, 2022. [Online]. Available:
  \url{https://github.com/ayman-elgharabawy/PNN}
\BIBentrySTDinterwordspacing

\bibitem{10.1007/978-3-030-34879-3_2}
M.~S. Meena, P.~Singh, A.~Rana, D.~Mery, and M.~Prasad, ``A robust face
  recognition system for one sample problem,'' in \emph{Image and Video
  Technology}, C.~Lee, Z.~Su, and A.~Sugimoto, Eds.\hskip 1em plus 0.5em minus
  0.4em\relax Cham: Springer International Publishing, 2019, pp. 13--26.

\bibitem{8615011}
S.~Rajora, D.~kumar Vishwakarma, K.~Singh, and M.~Prasad, ``Csgi: A deep
  learning based approach for marijuana leaves strain classification,'' in
  \emph{2018 IEEE 9th Annual Information Technology, Electronics and Mobile
  Communication Conference (IEMCON)}, 2018, pp. 209--214.

\bibitem{8258754}
A.~A. Padmanabha, M.~A. Appaji, M.~Prasad, H.~Lu, and S.~Joshi,
  ``Classification of diabetic retinopathy using textural features in retinal
  color fundus image,'' in \emph{2017 12th International Conference on
  Intelligent Systems and Knowledge Engineering (ISKE)}, 2017, pp. 1--5.

\bibitem{Wolfram}
\BIBentryALTinterwordspacing
{Wolfram Research, Inc., Mathematica}, ``Wolfram.'' [Online]. Available:
  \url{https://www.wolfram.com/mathematica/}
\BIBentrySTDinterwordspacing

\end{thebibliography}
